\documentclass[pdflatex,sn-mathphys-num]{sn-jnl}%

\usepackage{graphicx}%
\usepackage{multirow}%
\usepackage{amsmath,amssymb,amsfonts}%
\usepackage{amsthm}%
\usepackage{mathrsfs}%
\usepackage[title]{appendix}%
\usepackage{xcolor}%
\usepackage{textcomp}%
\usepackage{manyfoot}%
\usepackage{booktabs}%
\usepackage{algorithm}%
\usepackage{algorithmicx}%
\usepackage{algpseudocode}%
\usepackage{listings}%

\theoremstyle{thmstyleone}%

\theoremstyle{thmstyletwo}%

\theoremstyle{thmstylethree}%

\begin{document}

\title[Article Title]{When uncertainty quantification fails: structural vulnerabilities and corrections for selective prediction under clinical class imbalance}

\author*[1,2]{\fnm{Walter} \sur{Endrizzi}}\email{wendrizzi@fbk.eu}
\author[1]{\fnm{Flavio} \sur{Ragni}}\email{fragni@fbk.eu}
\author[1,3]{\fnm{Raffaele} \sur{Marchesi}}\email{rmarchesi@fbk.eu}
\author[1]{\fnm{Stefano} \sur{Bovo}}\email{sbovo@fbk.eu}
\author[1]{\fnm{Monica} \sur{Moroni}}\email{mmoroni@fbk.eu}
\author[1,4]{\fnm{Giuseppe} \sur{Jurman}}\email{jurman@fbk.eu}
\author[5]{\fnm{Venet} \sur{Osmani}}\email{v.osmani@qmul.ac.uk}

\affil[1]{\orgdiv{Data Science for Health}, \orgname{Fondazione Bruno Kessler}, \orgaddress{\city{Trento}, \country{Italy}}}

\affil[2]{\orgdiv{Department of Cellular, Computational and Integrative Biology}, \orgname{University of Trento}, \orgaddress{\city{Trento}, \country{Italy}}}

\affil[3]{Department of Mathematics, University of Pavia, Pavia, Italy}

\affil[4]{\orgdiv{Department of Biomedical Sciences}, \orgname{Humanitas University}, \orgaddress{\city{Milan}, \country{Italy}}}

\affil[5]{\orgdiv{Digital Environment Research Institute}, \orgname{Queen Mary University of London}, \orgaddress{\city{London}, \postcode{E1 1HH}, \country{United Kingdom}}}

\abstract{\textbf{Background.} Selective prediction systems improve the safety of clinical artificial intelligence by deferring ambiguous cases to human experts while automating
high-confidence decisions. However, this safety mechanism relies entirely on the
premise that uncertainty scores accurately reflect clinical ambiguity. In prognostic tasks characterized by severe class imbalance, where the clinical outcome of interest is a minority, this assumption breaks down. Standard uncertainty measures cease to capture clinically aligned ambiguity; instead, they become a proxy for class membership, systematically penalizing minority patients regardless of their individual predictive clarity.

\textbf{Methods.} We characterize three structural failure modes driving this breakdown: the mismatch between standard entropy's mathematical peak and the clinical decision threshold; the asymmetric spatial penalty in margin-based uncertainty; and structural uncertainty divergence of minority predictions driven by data scarcity. We propose three counter-mechanisms: an empirically validated \textit{threshold-aware ambiguity score} that realigns maximum uncertainty with the clinical threshold, a \textit{class-conditioned rejection} (CCR) mechanism ranking patients within their predicted trajectory rather than globally, and a class-stratified evaluation metric that prevents global risk scores from concealing automated minority exclusion.

\textbf{Results.} Across 56 synthetic conditions and three neurodegenerative cohorts spanning severe class imbalance to near-balanced conditions (multiple sclerosis, Parkinson's Disease, Alzheimer's Disease), our framework consistently maintained bounded, balanced risk across trajectories. In contrast, global metrics either collapsed minority Sensitivity or protected the minority at the expense of majority patients. An ablation study confirmed that applying CCR with standard entropy fails to protect the minority because it breaks within-class monotonicity when $\tau \neq 0.5$. Combining CCR with threshold-aware metrics restores strict monotonicity, correctly identifies ambiguous patients, and safely mitigates minority exclusion.

\textbf{Conclusions.} Under clinical class imbalance, standard uncertainty quantification measures dataset sparsity rather than clinical ambiguity, creating a systemic illusion of safety. This bias is a correctable structural flaw. Clinical pipelines deploying selective prediction under imbalance should adopt threshold-aligned uncertainty measures and class-stratified evaluation. This reduces class-specific risk imbalance, supporting safer selective prediction under the evaluated conditions without silently excluding the most vulnerable patients.}

\keywords{uncertainty, selective prediction, sample rejection, clinical AI, class imbalance, neurodegenerative disease, entropy, calibration, generalized risk-coverage curve}

\maketitle

\section{Introduction}\label{sec1}

Deployment of machine learning models in high-stakes clinical environments requires algorithmic systems that can reliably estimate their own limitations. In prognostic tasks, such as predicting the trajectories of neurodegenerative diseases \cite{poretto2025machine, cabral2015predicting, malaguti2025machine}, biological heterogeneity often makes definitive classification impossible. In these settings, the safety of an automated system depends on selective prediction: the mechanism by which an algorithm defers ambiguous cases to a human expert while automating high-confidence decisions \cite{lopez2025uncertainty}.

Standard selective prediction relies on uncertainty quantification (UQ) metrics to rank patients for deferral. However, in real-world clinical prognosis, datasets are often highly imbalanced, with the class of highest clinical interest (such as rapid progressors) representing only a small minority. Under these conditions, standard UQ approaches may fail to provide reliable estimates of true predictive uncertainty; instead, they can become proxies for class membership. Because minority patients inherently lack dense feature representations, current uncertainty estimates become structurally biased \cite{jimenez2026position, bickford2025rethinking}. Therefore, algorithms learn to equate the minority class with high uncertainty, failing to disentangle triage-relevant ambiguity from dataset sparsity \cite{dejong2024how, chen2023transfer}.

This measurement failure is not a consequence of poor model training, but rather the result of distinct structural vulnerabilities inherent to standard post-hoc UQ metrics as they are commonly deployed in applied heuristic pipelines. The first vulnerability, the threshold-entropy mismatch, occurs because the mathematical peak of standard entropy (i.e. Shannon entropy) is rigidly anchored to a perfectly balanced decision threshold ($0.50$). This structural misalignment with the naturally shifted thresholds ($\tau$) required for imbalanced data raises concerns about its appropriateness as an uncertainty measure \cite{wimmer2023quantifying}: when predicted probabilities fall near the balanced decision point, standard entropy may produce biased uncertainty estimates that reflect class imbalance rather than clinically aligned ambiguity. Consequently, clinically relevant minority patients are either erroneously automated or unnecessarily deferred due to an uncertainty calibration structure that is poorly aligned with the imbalanced decision space.

Although probability rescaling and threshold moving are well-established techniques for optimizing classification performance under severe imbalance \cite{collell2018simple}, their implications for realigning uncertainty estimation remain largely unexplored.

A second vulnerability concerns the spatial penalty introduced by margin-based uncertainty. The Classification Margin partially addresses the entropy-threshold mismatch by measuring uncertainty relative to the operative decision threshold rather than to $0.50$. However, it still penalizes predictions strictly by their absolute distance from the threshold. When the decision boundary is tightly compressed against the majority-class space, this can disproportionately flag confident majority-class patients for rejection. Margin-based UQ may appear to protect minority predictions, but it does so by systematically deferring confident majority-class cases, yielding an undesirable and clinically fragile trade-off. Although previous works have identified structural failure modes in margin-based confidence \cite{jiang2018trust} and learning with rejection \cite{cortes2016learning}, these analyses assume relatively balanced decision spaces and do not account for the asymmetric spatial penalty introduced by extreme class imbalance.

Finally, this ranking failure is compounded by structural uncertainty divergence. Because of their underlying sparsity, minority predictions inherently carry systematically higher ensemble disagreement than majority predictions, even when evaluated at the exact same distance from the decision boundary. Recent research in long-tailed medical imaging corroborates this phenomenon, observing that rare classes naturally exhibit wider confidence intervals strictly as a function of their frequency \cite{rezaei2023bayesian}. 

Taken together, these mechanisms show that \textbf{standard global ranking instruments} are intrinsically limited under severe class imbalance, as they can \textbf{conflate decision-level uncertainty with class prevalence, decision-boundary geometry, and minority-class sparsity}.

Standard evaluation frameworks commonly fail to detect these vulnerabilities. Although recent literature has begun to identify the structural limitations of standard selective evaluation paradigms \cite{pugnana2023auc}, the specific masking effect caused by extreme class imbalance remains unaddressed. As systems defer patients, safety is commonly evaluated using multi-threshold metrics such as the Area Under the Risk-Coverage Curve (AURC) \cite{geifman2017selective} or its recent refinement, the Area Under the Generalized Risk-Coverage Curve (AUGRC) \cite{traub2024overcoming}. However, because these metrics aggregate risk across the entire global population, an algorithm can mathematically improve its global score by systematically excluding the harder-to-predict minority class. In this way, standard evaluation can reward automated minority exclusion, masking algorithmic bias under the guise of improved overall safety.

\textbf{Contributions}. Addressing this gap requires moving beyond global ranking and toward class-stratified uncertainty measurement and evaluation. In this paper, we characterize the structural vulnerabilities of standard UQ under clinical class imbalance, propose a framework of well-motivated, empirically validated measurement and evaluation instruments, and demonstrate their efficacy across synthetic and real-world neurodegenerative prognostic tasks. We validate the proposed framework across 56 controlled synthetic conditions and three real-world neurodegenerative patient cohorts comprising multiple sclerosis, Parkinson’s Disease, and Alzheimer’s Disease. Specifically, we make the following contributions:
\begin{itemize}
    \item \textbf{Threshold-aware ambiguity score ($H_\tau$):} We demonstrate that standard Shannon entropy fails under class-conditioned rejection (CCR) when $\tau \neq 0.5$ by breaking within-class monotonicity. To restore correct ordinal ranking, we introduce $H_\tau$, an empirically validated geometric construction. By design, $H_\tau$ operates as a non-linear, information-theoretic alternative to the linear decision margin, satisfying three key structural properties: continuity across the probability space, strict monotonicity within each predicted class, and exact reduction to standard Shannon entropy at parity ($\tau = 0.5$).
    \item \textbf{CCR (Class-Conditioned Rejection):} A class-conditioned rejection mechanism that isolates intra-class ambiguity, establishing a structural, mechanical mechanism to prevent minority exclusion.
    \item \textbf{ccAUGRC (Class-Conditioned Area Under the Generalized Risk-Coverage Curve):} An evaluation decomposition that prevents risk-masking by quantifying the generalized risk-coverage trade-off independently for each clinical trajectory.
\end{itemize}

\subsection{Related Work}

\textbf{Selective prediction foundations.} The foundations of selective prediction have been extensively formalized to optimize the trade-off between coverage and accuracy \cite{chow1970optimum, elyaniv2010foundations, gangrade2021selective}. While these seminal theoretical frameworks natively accommodated the decision threshold as a free parameter, modern applied pipelines often take a more rigid approach. Contemporary systems successfully leverage ensemble-based or Bayesian methods to generate predictive distributions for deferral \cite{delcorso2025shedding}, but they frequently default to unaligned ranking metrics such as standard Shannon entropy. Because this post-hoc ranking predominantly assumes balanced class priors, and because disentangling aleatoric from epistemic uncertainty remains an open challenge \cite{mucsanyi2024benchmarking, bindini2026dealing}, standard selective classification frequently magnifies performance disparities across distinct subgroups \cite{jones2021selective, lópez2026empirical}. This vulnerability underscores the critical need for class-aware triage strategies.

\textbf{Entropy limitations and threshold-awareness.} While base-classifier training frequently addresses imbalance via loss-weighting or resampling \cite{chawla2002smote, lin2017focal}, and recent approaches integrate learning to reject directly with long-tail learning paradigms \cite{narasimhan2024learning}, the calibration of post-hoc uncertainty under imbalance remains critically under-explored. Existing solutions largely focus on probabilistic recalibration or multimodal data acquisition \cite{hoarau2026reducing}, but lack structural geometric corrections for the threshold-entropy mismatch and the asymmetric spatial penalties introduced by shifted decision boundaries, as identified in recent theoretical work \cite{wimmer2023quantifying, jimenez2026position}.

\textbf{Evaluation corrections.} The Generalized Risk-Coverage curve and AUGRC have become standard for evaluating selective models across dynamic operating points \cite{traub2024overcoming}. While methods exist to improve multi-threshold metrics, and concurrent work has recently introduced class-averaged risk-coverage measures using empirical quantiles for multiclass settings \cite{SAGLAM2026105084}, these approaches primarily address triage routing rather than the underlying signal geometry. The literature requires a dedicated decomposition of global risk-coverage to account for the specific risk displacement artifacts caused by extreme clinical class imbalance, a gap this work addresses via ccAUGRC.

\textbf{Conformal prediction and risk control.} An alternative, prominent approach to managing uncertainty under imbalance is conformal prediction (CP) and distribution-free risk control \cite{angelopoulos2021gentle}. Class-conditional conformal prediction (CCCP) explicitly addresses the problem of bounding risk within subpopulations by generating adaptive prediction sets that satisfy a user-defined coverage guarantee for each class independently \cite{sadinle2019least, ding2023class}. While CCCP and distribution-free risk control represent the standard for theoretical coverage guarantees, they solve a fundamentally different operational problem (bounding error rates via variable-sized prediction sets) than selective classification (managing a fixed clinical deferral bandwidth via fixed rejection quotas).

\section{Methods}\label{sec2}
\subsection{Problem formulation}\label{sec2:problem}
Let $\mathcal{D} = \{(x_i, y_i)\}_{i=1}^{N}$ denote a binary clinical prognostic dataset where $y_i \in \{0, 1\}$ indicates the absence or presence of disease progression. In imbalanced clinical settings, the class prior $\pi = P(y=1)$ is substantially less than 0.5, reflecting the natural prevalence of the prognostic outcome in the population. A probabilistic classifier $f\colon \mathcal{X} \rightarrow [0,1]$ produces a calibrated predicted probability $p_i = f(x_i)$. Clinical decisions are made using a decision threshold $\tau \in (0,1)$ derived from the population prior \cite{vickers2006decision, elkan2001foundations}: a patient is predicted as a progressor if $p_i \geq \tau$ and as stable otherwise. In standard practice, $\tau = \pi$, ensuring that the classifier's operating point is aligned with the clinical prevalence of the outcome rather than the mathematical center of the probability space.

A \textit{selective prediction system} augments this classifier with a rejection function $g\colon \mathcal{X} \rightarrow \{0,1\}$, where $g(x_i) = 0$ defers patient $i$ to a clinician for manual review. Deferral is operationalized by thresholding an uncertainty score $u_i$: predictions exceeding a predefined limit are deferred to human experts. The retained cohort $\mathcal{A} = \{i : g(x_i) = 1\}$ is processed automatically; the deferred cohort $\mathcal{R} = \{i : g(x_i) = 0\}$ receives human review.

The central premise of selective prediction is that the uncertainty score $u_i$ correctly identifies algorithmic ambiguity relative to the clinical decision threshold. In the following subsections, we formalize three structural vulnerabilities where this premise breaks down under clinical class imbalance, and subsequently define the algorithmic framework proposed to address them.
 
\subsection{Structural vulnerabilities of standard uncertainty quantification}  
\begin{figure}[htbp]
    \centering
    \includegraphics[width=\textwidth]{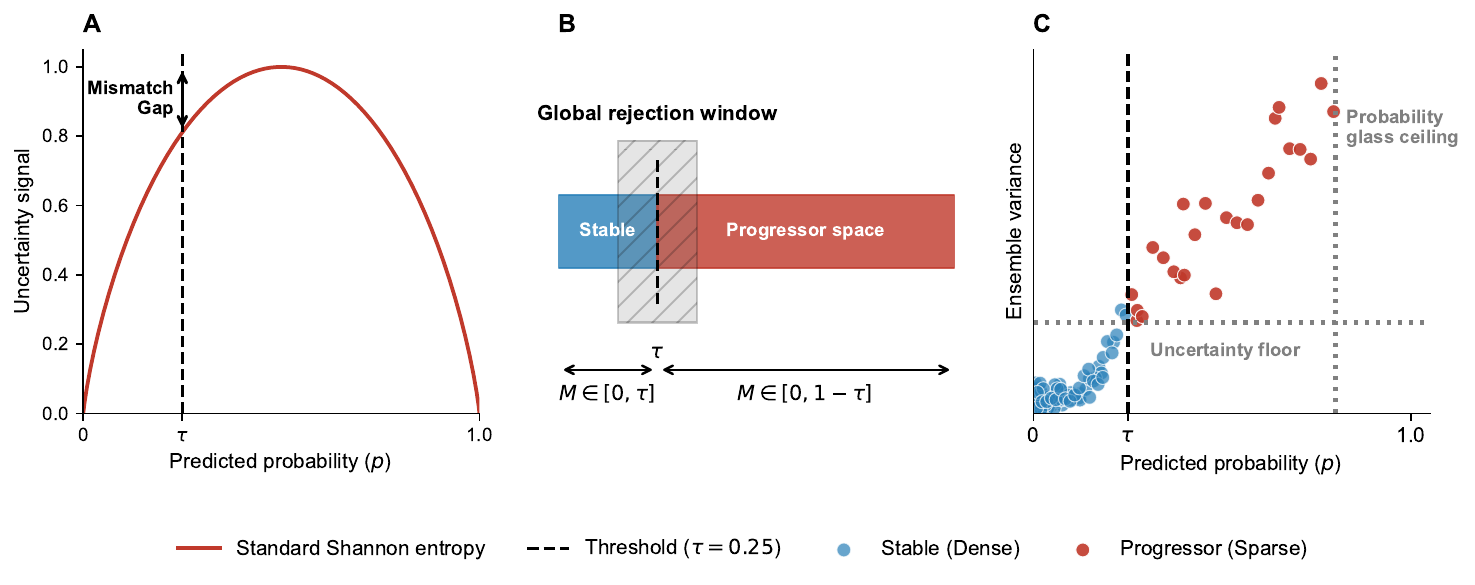}
    \caption{\textbf{Structural vulnerabilities of standard uncertainty quantification under clinical class imbalance.} The three panels conceptualize the mathematical failure modes of global uncertainty metrics in a binary classification setting when the clinical decision threshold ($\tau$) is shifted from 0.50. 
    \textbf{(A) Threshold-entropy mismatch.} Standard Shannon entropy (red curve) is mathematically anchored to a perfectly balanced decision space. Evaluated at a naturally shifted clinical threshold (e.g., $\tau=0.15$), a critical mismatch gap emerges. Maximum mathematical entropy fails to align with maximum clinical ambiguity ($p=\tau$), causing standard entropy to operate as a distance metric from 0.5 rather than a true measure of clinical uncertainty. 
    \textbf{(B) The spatial penalty.} Anchoring uncertainty to the absolute decision margin $M(p) = |p-\tau|$ exposes the asymmetric geometry of the probability space. The majority class is structurally compressed into a narrow margin ($M \in [0, \tau]$), whereas the minority class lies in a vastly larger space ($M \in [0, 1-\tau]$). Applying a single global rejection window systematically distorts the retained cohort by disproportionately penalizing or entirely excluding the compressed majority class. 
    \textbf{(C) Structural uncertainty divergence.} An idealized illustration demonstrating the structural vulnerabilities introduced by extreme class imbalance. A combination of aleatoric noise and epistemic sparsity compresses minority predictions (red) into a bounded space, imposing a theoretical probability glass ceiling ($g < 1$, dotted line). Forced into this compressed regime, the minority class theoretically exhibits a structural uncertainty floor (dashed line), meaning their ensemble disagreement remains artificially elevated above majority patients (blue) even at identical mathematical distances from the decision boundary ($\tau$, solid line). We empirically validate the real-world manifestation of this structural asymmetry in Section \ref{sec3}.}
    \label{fig:structural_vulnerabilities}
\end{figure}

We characterize three distinct mathematical failure modes that arise when standard uncertainty metrics are applied to imbalanced clinical prognostic tasks, which are conceptually illustrated in Figure \ref{fig:structural_vulnerabilities}.

\subsubsection{Vulnerability 1. Threshold-entropy mismatch}
The most widely used uncertainty metric in selective prediction is predictive entropy, which in binary classification reduces to Shannon binary entropy \cite{gawlikowski2021survey, smith2018understanding}:
\begin{equation}
  H(p) \;=\; -p \log_2 p \;-\; (1-p)\log_2(1-p).
\end{equation}
A fundamental limitation of $H(p)$ is that it achieves its global maximum exclusively at $p = 0.5$, independently of the clinical decision threshold $\tau$. This follows directly from setting the derivative to zero: $\frac{dH}{dp} = -\log_2 p + \log_2(1-p) = 0$ only when $p = 0.5$, confirming a unique global maximum at the mathematical center of the probability space.
 
Consequently, a patient predicted exactly at the decision boundary $p = \tau$, the point of maximum clinical ambiguity, achieves an entropy $H(\tau) < H(0.5) = 1$ for any $\tau \neq 0.5$. Conversely, a patient confidently predicted at $p = 0.5$ reaches maximum entropy regardless of $\tau$. For example, in a disease with prevalence $\pi = 0.10$, confident progressor predictions near $p = 0.50$ yield maximal entropy. Under these conditions, standard entropy operates as a distance metric from $p = 0.5$ rather than a measure of triage-relevant ambiguity, disproportionately flagging minority patients for deferral. This vulnerability is naturally instantiated by standard global Shannon entropy ($H_{\text{Total}}$).
 
\subsubsection{Vulnerability 2. Spatial penalty}
A natural correction for Vulnerability~1 is to replace entropy with the absolute decision margin $M(p) = |p - \tau|$, which anchors uncertainty to the clinical threshold $\tau$ rather than to $0.5$. This eliminates the entropy mismatch, but introduces a distinct structural anomaly arising from the asymmetric geometry of the probability space around $\tau$.
 
For patients predicted stable ($p < \tau$), the margin is strictly bounded within $[0,\, \tau]$: the most certain stable patient achieves at most $M = \tau$. For patients predicted as progressors ($p > \tau$), the margin spans $[0,\, 1-\tau]$: the most certain progressor can
achieve $M = 1-\tau$. When $\tau < 0.5$ the two classes occupy
structurally different certainty spaces. With $\tau = 0.10$, for
instance, the entire stable class is compressed within
$M \in [0,\, 0.10]$, while the progressor class has access to
$M \in [0,\, 0.90]$.
 
While unequal margin supports do not strictly guarantee unequal retention, as this ultimately depends on the density of the class-conditional distributions—under adequate class separability, this asymmetry makes it impossible to find a single global threshold $\delta$ that fairly partitions both classes. If $\delta > \tau$, no stable patient can achieve $M > \delta$, because their entire margin space lies below $\delta$, leading to deferral of the entire stable class (majority patients). If $\delta \leq \tau$, some stable patients can be retained, but the threshold simultaneously retains every progressor whose margin exceeds $\delta$, since their margin space is far larger. In either regime, the retained cohort is systematically distorted: the stable class either disappears entirely or is underrepresented relative to its clinical prevalence. This pathology is naturally instantiated by global margin thresholding.

\subsubsection{Vulnerability 3. Structural uncertainty divergence}
Even if an uncertainty metric is perfectly anchored to the decision threshold $\tau$, a third structural vulnerability persists: structural uncertainty divergence. This phenomenon causes global uncertainty rankings to systematically defer minority patients through two compounding effects. 

First, minority predicted probabilities are compressed into a restricted range $[\tau, g]$ with a structural upper bound $g < 1$, which we term the probability glass ceiling. Two distinct factors drive this ceiling. Biological heterogeneity in the minority class introduces genuine, irreducible aleatoric uncertainty, which inherently limits prediction confidence. Concurrently, data sparsity prevents the model from resolving confident minority-class boundaries, driving up ensemble disagreement (often used as a proxy for epistemic uncertainty). This sparsity-driven uncertainty is further amplified by standard machine learning pipeline mechanics. For instance, ensemble aggregation inherently smooths extreme probabilities. Furthermore, post-hoc probability calibration (such as Platt scaling \cite{platt1999probabilistic}) fitted on imbalanced data heavily weights the majority prior; consequently, the calibrator acts conservatively, systematically penalizing and pulling down high-confidence minority predictions. Together, these factors artificially compress the minority probability range into a glass ceiling, preventing the model from assigning high-confidence minority predictions regardless of how clear the patient's actual prognosis might be.

Second, because minority predictions are constrained to a compressed range $[\tau,g]$ rather than covering the entire interval $[\tau, 1]$, they are functionally compressed relative to majority predictions, which the model can confidently assign to the stable domain near $p \approx 0$. This creates an uncertainty floor: even the most certain minority prediction carries more absolute uncertainty than a typical majority prediction. This elevation is a modeling artifact rather than a reflection of the patient's individual clinical ambiguity: it arises from the model's data limitations and is shared by all minority patients regardless of their true prognostic clarity. Ultimately, due to this floor, the global ranking by any absolute uncertainty score preferentially rejects minority patients, as their absolute scores remain structurally elevated above those of majority patients at equivalent clinical distances from the decision threshold.
 
\subsection{Proposed framework}
Because the three vulnerabilities stem from distinct layers of the  selective prediction pipeline, i.e. the metric formulation, the probability  geometry, and the learning dynamics, they cannot be resolved by a single algorithmic adjustment. Instead, we propose a cohesive framework that intervenes sequentially: first at the signal layer to correct the metric  formulation, second at the triage layer to correct the sorting geometry,  and finally at the evaluation layer to accurately measure clinical risk.
 
\subsubsection{Threshold-aware ambiguity score}\label{sec:htau}
To correct the entropy mismatch (Vulnerability~1) at the signal layer, we propose $H_\tau$, constructed by applying a piecewise linear rescaling  of the predicted probability that maps the clinical decision boundary to the entropy peak:
\begin{equation}
  \tilde{p} \;=\;
  \begin{cases}
    \dfrac{p}{2\tau}
      & \text{if } 0 \leq p \leq \tau, \\[10pt]
    0.5 + \dfrac{p - \tau}{2(1-\tau)}
      & \text{if } \tau < p \leq 1.
  \end{cases}
  \label{eq:rescaling}
\end{equation}
The final threshold-aware ambiguity score is then computed as:
\begin{equation}
  H_\tau(p) \;=\;
  -\tilde{p}\,\log_2 \tilde{p}
  \;-\; (1-\tilde{p})\,\log_2(1-\tilde{p}).
  \label{eq:htau}
\end{equation}
 
The transformation $p \mapsto \tilde{p}$ maps $[0, \tau]$ to $[0, 0.5]$ and $(\tau, 1]$ to $(0.5, 1]$ monotonically, ensuring $\tilde{p}(\tau) = 0.5$. Consequently, $H_\tau(p)$ attains its maximum strictly at the clinical boundary $p = \tau$. The metric possesses two core properties: first, it reduces to standard Shannon entropy when $\tau = 0.5$ ($H_\tau \equiv H$), acting as a strict generalization. Second, the rescaling is monotone-preserving within each predicted class, ensuring patients closer to $\tau$ receive higher $H_\tau$ scores than patients further from $\tau$ within the same trajectory. The mapping is continuous at $p = \tau$ but not differentiable there unless $\tau = 0.5$: the left-hand derivative is $1/(2\tau)$ and the right-hand derivative is $1/(2(1-\tau))$, producing a sharp peak at the decision boundary for any $\tau \neq 0.5$. Because the framework uses ordinal ranking of $H_\tau$ scores this non-smoothness does not affect the algorithm's operation.

Due to this strict within-class monotonicity, any rejection mechanism applied independently to each side of the threshold will induce the exact same ordinal ranking as a simple linear margin ($\vert{}p - \tau\vert{}$). Despite this operational equivalence for triage, we utilize $H_\tau$ as our primary signal for its non-linear Sensitivity and information-theoretic interpretability. Unlike a linear margin which evaluates probability shifts linearly across the entire probability space, the logarithmic nature of $H_\tau$ penalizes predictions more severely when they fall near the operational threshold $\tau$. This provides a probabilistically bounded measure of missing information that aligns with standard information-theoretic definitions of uncertainty.

$H_\tau$ resolves Vulnerability~1 by anchoring maximum ambiguity to the clinical decision boundary. Applied globally, however, it does not resolve Vulnerability~3: because epistemic sparsity restricts minority predictions to a compressed probability range, their absolute $H_\tau$ scores are subject to an artificially elevated uncertainty floor. Consequently, even at comparable clinical distances from $\tau$, minority patients remain structurally penalized under a global ranking, preferentially triggering their deferral.  This motivates the class-conditioned rejection mechanism described in the following section.

\subsubsection{Class-Conditioned Rejection}
To address the spatial penalty (Vulnerability 2) and structural uncertainty divergence  (Vulnerability 3), we propose CCR (Class-Conditioned Rejection). CCR partitions the cohort by predicted class and ranks patients by uncertainty within each partition independently (Algorithm \ref{algo:CCR}).
 
\begin{algorithm}
\caption{Class-Conditioned Rejection (CCR)}\label{algo:CCR}
\begin{algorithmic}[1]
\Require Cohort $\mathcal{D}$, predicted probabilities $p$, uncertainty signal $u$, rejection rate $r \in [0, 1)$, decision threshold $\tau$
\Ensure Retained cohort $\mathcal{A}$ 
\State $\mathcal{D}_0 \gets \{i \in \mathcal{D} \mid p_i < \tau\}$ \Comment{Partition predicted stable}
\State $\mathcal{D}_1 \gets \{i \in \mathcal{D} \mid p_i \geq \tau\}$ \Comment{Partition predicted progressor}
\State Sort $\mathcal{D}_0, \mathcal{D}_1$ in descending order by uncertainty signal $u$
\State $k_0 \gets \lfloor r \times |\mathcal{D}_0| \rfloor$ \Comment{Rejection quota for stable class}
\State $k_1 \gets \lfloor r \times |\mathcal{D}_1| \rfloor$ \Comment{Rejection quota for progressor class}
\State $\mathcal{R}_0 \gets$ top $k_0$ patients from $\mathcal{D}_0$
\State $\mathcal{R}_1 \gets$ top $k_1$ patients from $\mathcal{D}_1$
\State $\mathcal{A} \gets \mathcal{D} \setminus (\mathcal{R}_0 \cup \mathcal{R}_1)$ \Comment{Merge remaining patients}
\State \textbf{return} $\mathcal{A}$
\end{algorithmic}
\end{algorithm}
 
By rejecting fraction $r$ from each predicted class independently, CCR ensures the preservation of the predicted class proportions (up to a finite-sample rounding error) in the retained cohort. While this structural class-preservation prevents the systematic depletion of a predicted trajectory, it is equally important to examine how this mechanism filters misclassifications (false negatives and false positives) within these isolated queues. 

In the $\mathcal{D}_0$ queue (predicted stable), false negatives typically lie near the decision boundary ($p \approx \tau$). Because all probabilities in $\mathcal{D}_0$ are strictly bounded below $\tau$, even standard entropy increases monotonically toward this boundary. Consequently, simply isolating the $\mathcal{D}_0$ queue forces both standard and threshold-aware metrics to correctly assign the highest uncertainty to boundary false negatives, pushing them to the front of the rejection list.

Conversely, reliable ranking in the $\mathcal{D}_1$ queue (predicted progressor) mathematically necessitates a threshold-aware metric. As established, when $\tau < 0.5$, the mathematical peak of standard entropy ($0.5$) falls inside the $\mathcal{D}_1$ space. Applying standard entropy here breaks within-class monotonicity, mistakenly assigning maximum uncertainty to confident predictions near $p \approx 0.5$ rather than to boundary false positives. Threshold-aware metrics, such as $H_\tau$ or the decision margin, resolve this by anchoring peak uncertainty exactly at $\tau$. Thus, while CCR provides the macro-level mechanism for class preservation, it must be paired with a $\tau$-centered metric to correctly filter boundary misclassifications across both ordinal queues.
Furthermore, by ranking ordinally within the predicted trajectory, CCR bypasses the uncertainty floor: minority patients are compared only to other minority patients, neutralizing their structurally elevated absolute uncertainty.

\subsubsection{Class-conditioned Area Under the Generalized Risk-Coverage Curve (ccAUGRC)}
Standard evaluation metrics for selective prediction compute performance globally across all patients. Under class imbalance, this aggregation introduces a structural evaluation artifact caused by inter-class substitution: a triage mechanism can artificially minimize its overall risk by systematically replacing minority patients (whose epistemic compression, Vulnerability~3, produces structurally higher error rates) with lower-risk majority patients in the accepted cohort. Global evaluation is therefore blind to whether safety is actually maintained across distinct patient trajectories.

To expose and resolve this substitution artifact, we adapt the recently formalized Area Under the Generalized Risk-Coverage curve (AUGRC;~\cite{traub2024overcoming}). While standard AUGRC defines risk at a global coverage $c$ as the fraction of total errors in the accepted cohort over the original population $N$, we compute this trade-off strictly within each class subpopulation $k \in \{0,1\}$. Let $c_k \in [0,1]$ denote the fraction of class-$k$ patients retained at a given rejection rate $r$, and $E_k$ the corresponding number of misclassified class-$k$ patients. The class-conditioned AUGRC (ccAUGRC) is defined as:
\begin{equation}
  \mathrm{ccAUGRC}_k
  \;=\;
  \int_{c_k(r_{\max})}^{1}
  \frac{E_k}{N_k}
  \;\mathrm{d}c_k,
  \label{eq:ccaugrc}
\end{equation}
where $N_k$ is the total number of class-$k$ patients. The formulation is structurally identical to AUGRC but substitutes the global population denominator $N$ with the class-specific denominator $N_k$, and the global coverage $c$ with the class-specific coverage $c_k$. The integral is approximated by the trapezoidal rule over the discrete pairs $\{(c_k(r),\, E_k(r)/N_k)\}_{r \in [0,\, r_{\max}]}$, sorted by increasing $c_k$; in all experiments $r_{\max} = 0.80$.

ccAUGRC is not a new metric but a class-stratified application of AUGRC using the same integration machinery. Reporting ccAUGRC for both classes alongside global AUGRC allows a system that achieves low global AUGRC through automated minority exclusion to be distinguished from one that maintains bounded risk across both trajectories simultaneously.
 
\subsection{Datasets}
\subsubsection{Real-World clinical cohorts}
We validate our framework on three neurodegenerative disease cohorts spanning distinct imbalance regimes. Each cohort represents an independent prognostic task with disease-specific features, clinical endpoints, and population characteristics. The intentional heterogeneity across feature spaces and clinical endpoints is a design choice: we are not benchmarking a classifier but validating an evaluation framework, which should generalize across task characteristics and vary predictably with the class prior $\pi$. In all three cohorts the decision threshold $\tau$ is set equal to the observed class prior $\pi$, consistent with standard clinical practice and with the problem formulation in Section~\ref{sec2:problem}.

\begin{itemize}
    \item \textbf{Multiple sclerosis (MS, severe imbalance, $\pi=0.13$).} $N=719$ patients from a multicentric Italian cohort described in full detail in Poretto et al.~\cite{poretto2025machine}. Briefly, patients were recruited from three Italian MS centers (Careggi University Hospital, Florence; IRCCS Ospedale Policlinico San Martino, Genoa; ASUIT, Trento) and diagnosed according to the 2010 McDonald criteria. The outcome variable is Progression Independent of Relapse Activity (PIRA) at year 3, defined as EDSS worsening in the absence of relapse activity~\cite{muller2023harmonizing}. Minority class: progressors ($N=92$, 12.8\%). Predictive features cover demographic, clinical, and MRI-derived measures collected at baseline and two-year follow-up. Ethics approval and full preprocessing details are reported in~\cite{poretto2025machine}.

    \item \textbf{Parkinson's Disease (PD, moderate imbalance, $\pi=0.39$).} $N=247$ patients from a multicentric Italian cohort (NeuroArtP3 project) described in full detail in Endrizzi et al.~\cite{Endrizzi2026Explainable}. Briefly, patients were recruited from two Northern Italian sites (APSS, Trento; IRCCS San Martino Hospital, Genoa) and diagnosed according to standard clinical criteria. The outcome variable is the presence of Levodopa-induced dyskinesia (LID) within three years from the baseline visit. For this analysis, the predictive task focuses on forecasting future dyskinesia status independently of the patients' baseline motor complications. Minority class: patients with future LID ($N=96$, 39.0\%). Predictive features encompass baseline clinical-demographic data, core motor and non-motor assessments (e.g., MDS-UPDRS III, Hoehn \& Yahr), and detailed medication history. Ethics approval and full preprocessing details are reported in~\cite{Endrizzi2026Explainable}. Note that while the baseline cohort comprises 247 patients, the internal stratification logic of the cross-validation pipeline drops one patient due to uneven group splits; consequently, all empirical evaluations accurately reflect the 246 patients evaluated in the test folds.

    \item \textbf{Alzheimer's Disease (AD, near-balanced, $\pi=0.54$).} $N=314$ patients from the Alzheimer's Disease Neuroimaging Initiative (ADNI; \url{adni.loni.usc.edu}). The predictive task is the conversion from Mild Cognitive Impairment (MCI) to Alzheimer's Disease within 36 months of baseline assessment. Class 1: MCI-to-AD converters ($N=170$, 54.1\%). Predictive features encompass baseline cognitive assessments (e.g., ADAS-Cog, MMSE) and demographic information. Complete details regarding cohort selection, inclusion/exclusion criteria, and data preprocessing for this specific analytic sample are provided in Appendix \ref{secA1}.
\end{itemize}

The three cohorts span $\pi \in \{0.13, 0.39, 0.54\}$, covering the severe, moderate, and near-balanced imbalance regimes. This range was selected to provide systematic coverage of the operating conditions under which the framework predicts distinct effects.

\subsubsection{Synthetic data}
Synthetic data are used to establish a controlled phase diagram of standard UQ failure as a function of class prior $\pi$ and class separability $d$, providing a reference frame against which the real-world results are interpreted.
While these synthetic distributions assume Gaussian normality to isolate mathematical vulnerabilities, the subsequent evaluation on real-world clinical cohorts serves to explicitly validate these failure modes under highly non-Gaussian, non-parametric conditions.

Each synthetic condition is defined by two parameters. The class prior $\pi \in \{0.05, 0.10, 0.15, 0.20, 0.30, 0.40, 0.50\}$ determines the proportion of minority patients and simultaneously serves as the decision threshold ($\tau$). The separability $d \in \{0.5, 1.0, 2.0, 3.0\}$ is the mean separation between classes in standard deviation units (Cohen's $d$), ranging from heavily overlapping to well-separated class distributions.

Features ($M=20$ per patient) are generated from class-conditional Gaussian distributions with an exponentially decaying signal across features:
\begin{equation}
x^{(j)} \sim 
\begin{cases} 
\mathcal{N}(0, 1) & \text{class } 0 \\ 
\mathcal{N}(d \cdot e^{-0.5j}, 1) & \text{class } 1 
\end{cases}
\end{equation}
where $j \in \{0, 1, \dots, M-1\}$ is the feature index. The exponential decay ensures that the classification signal is concentrated in a smaller subset of features, avoiding trivially separable tasks at high $d$. Two sample sizes are used ($N \in \{300, 1000\}$). The complete grid of $7 \times 4 \times 2 = 56$ conditions is processed through the same ensemble calibration pipeline described in the next sections.

\subsection{Experimental setup}
\subsubsection{Asynchronous ensemble pipeline}
\begin{figure}
    \centering
    \includegraphics[width=1\linewidth]{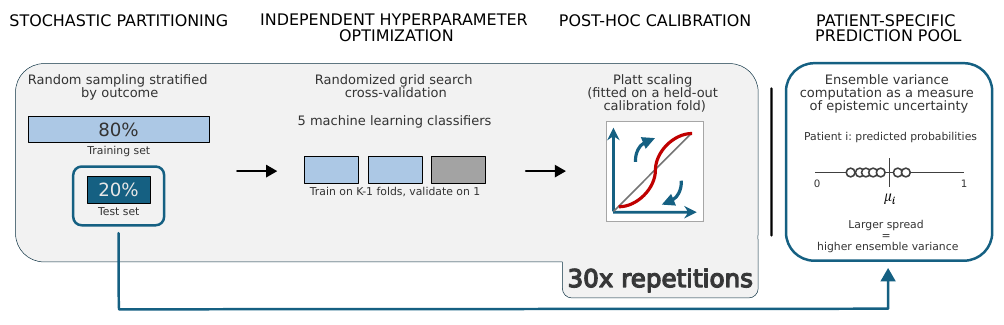}
    \caption{\textbf{Asynchronous ensemble pipeline.} For each of 30 bootstrap iterations, the dataset was split into stratified 80\% training and 20\% test partitions while preserving the natural class distribution. Within each iteration, five heterogeneous machine learning classifiers were independently optimized using randomized grid-search cross-validation and subsequently calibrated by Platt scaling on a held-out calibration fold to obtain reliable predicted probabilities. For each patient appearing in a test set, calibrated predictions across classifiers and iterations were aggregated into a patient-specific prediction pool. The mean prediction $\mu_i$ represents the ensemble consensus, while the variance $\sigma_i^2$ quantifies prediction variability across the ensemble and is used as a proxy for epistemic uncertainty. No external resampling strategy was applied, ensuring that the resulting uncertainty estimates reflect model behavior under the natural class imbalance.}
    \label{fig:methods}
\end{figure}

All experiments use a consistent asynchronous ensemble pipeline (Figure \ref{fig:methods}). The pipeline is identical across all three cohorts and all synthetic conditions, ensuring that performance differences reflect the uncertainty signal and rejection mechanism rather than modeling choices. This workflow is used to produce a single calibrated mean predicted probability for each patient who falls in the test set during model testing and can be decomposed into four macro stages.

\textbf{Stochastic partitioning.} For each of 30 bootstrap iterations, the dataset is split into 80\% training and 20\% test sets using stratified random sampling with a unique random seed per iteration, preserving the natural class distribution.

\textbf{Independent hyperparameter optimization.} Five heterogeneous base classifiers are trained independently per iteration: Gradient Boosted Trees (XGBoost), Random Forest, Extra Trees Classifier, Logistic Regression, and Support Vector Classifier with RBF kernel. Each classifier undergoes independent randomized grid search cross-validation ($n_{\text{iter}}=20$, 3-fold CV, 5 repeats) optimizing the Area Under the Precision-Recall Curve (AUPRC), chosen as the optimization target because it is sensitive to minority class performance and appropriate for imbalanced datasets.

\textbf{Post-hoc calibration.} To ensure predicted probabilities reflect reliable estimates of class membership, each trained classifier is independently calibrated using Platt scaling. To prevent data leakage, the scaling model is fitted on a held-out calibration fold within the same training iteration. This calibration is applied prior to ensemble aggregation to avoid score compression artifacts.

\textbf{Patient-specific prediction pool.} For each patient appearing in a test set, calibrated predictions from all models across all iterations in which that patient was in the test set are aggregated. The ensemble consensus $\mu_i$ (mean) and $\sigma_i$ (standard deviation) provide patient-specific estimates of the predicted probability and its variability across the ensemble, respectively. The ensemble disagreement $\sigma_i^2$ provides a measure of predictive uncertainty, which serves as a proxy for epistemic uncertainty and is used in the asymmetry test (Section \ref{sec3.2.1}).

No external data resampling techniques (e.g., SMOTE, random undersampling) were applied to alter the natural class distribution of the training folds. While algorithmic cost-sensitive learning (e.g., class-weighted loss functions) was included in the hyperparameter search space to aid minority class boundary detection, post-hoc Platt scaling ensures that the final probabilities remain calibrated to the natural prior $\tau$ regardless of the training-time weighting scheme. Complete information on the computational environment and hyperparameter optimization can be found in the Appendix \ref{secA2}.

\subsubsection{Baseline and Control Methods}
Six uncertainty quantification and rejection methods are compared across all experiments (Table~\ref{tab:methods}). $H_{\text{Total}}$ and Margin are established baselines. $H_\tau$ (Global) applies the proposed ambiguity score without the class-conditioned rejection mechanism. $H_{\text{Total}}$~+~CCR applies standard Shannon entropy within the class-conditioned framework, serving as a critical baseline to demonstrate the mathematical necessity of threshold-aware scoring when $\tau \neq 0.5$. $H_\tau$~+~CCR is the proposed dual architecture. Random~+~CCR is a control condition that replaces the $H_\tau$ signal with random within-class ordering while retaining the class-conditioned rejection mechanism, enabling decomposition of the two components' independent contributions.
Finally, we acknowledge Margin~+~CCR as a theoretical baseline, utilizing the linear decision margin ($|p_i - \tau|$) as the uncertainty signal within the CCR framework. As established in Section \ref{sec:htau}, both the margin and $H_\tau$ are strictly monotonic with respect to the distance from $\tau$ on either side of the decision boundary. Because Algorithm \ref{algo:CCR} relies exclusively on within-class ordinal ranking, applying Margin~+~CCR selects the exact same patients for rejection at every threshold as $H_\tau$~+~CCR, resulting in zero empirical disagreements. Consequently, the explicit Margin~+~CCR baseline produces an empirical rejection curve that perfectly overlaps with $H_\tau$~+~CCR. To prevent visual redundancy in the figures and duplicate rows in the performance tables, this baseline is omitted from the reported results, though we fully acknowledge its operational equivalence.
\begin{table}[ht]
\centering
\caption{\textbf{Uncertainty methods compared across all experiments.} Summary of the evaluated quantification and rejection strategies. The proposed architecture ($H_\tau$~+~CCR) is isolated against standard global baselines (Margin, $H_\text{Total}$), an ablation of the rejection mechanism ($H_{\tau}$), a standard entropy CCR baseline ($H_\text{Total}$~+~CCR) to demonstrate the necessity of threshold-aware scoring, and a randomized control (Random~+~CCR) to isolate the mechanical benefit of class preservation.}
\label{tab:methods}
\begin{tabular}{lll}
\hline
\textbf{Method} &
\textbf{Uncertainty signal} &
\textbf{Rejection} \\
\hline
$H_{\text{Total}}$ (Global)
  & Shannon entropy $H(p_i)$
  & Global \\
Margin (Global)
  & Decision margin $|p_i - \tau|$
  & Global \\
$H_\tau$ (Global)
  & Ambiguity Score $H_\tau(p_i)$
  & Global \\
  $H_{\text{Total}}$~+~CCR
  & Shannon entropy $H(p_i)$
  & Class-conditioned \\
$H_\tau$~+~CCR
  & Ambiguity Score $H_\tau(p_i)$
  & Class-conditioned \\
Random~+~CCR
  & Random
  & Class-conditioned \\
\hline
\end{tabular}
\end{table}

\subsubsection{Evaluation protocol and statistical analysis}
To capture the dynamic risk-coverage trade-off, all methods are evaluated across the full rejection rate range $r \in [0, 0.80]$ in steps of $0.05$. Performance is tracked via three class-stratified metrics: Sensitivity (the primary safety metric for progressive diseases), Specificity, and the Area Under the Precision-Recall Curve (AUPRC) in the retained cohort. 

To formally quantify these trajectories in the clinical cohorts, we compute three summary metrics. First, we calculate \textit{Sensitivity stability}, defined as the slope of target-class Sensitivity over the $0\%\text{--}50\%$ rejection range estimated via linear regression. Because the 0\%-50\% range captures the most plausible deployment scenarios, a negative slope here indicates active target-class exclusion, whereas a slope near zero indicates stable protection. Second, to empirically quantify the risk-masking effect identified in standard evaluations, we compute the class-conditioned AUGRC (\textit{ccAUGRC}) for both progressor and stable trajectories independently, integrated over the full rejection range. Finally, standard \textit{Global AUGRC} is reported to demonstrate how class-specific failures are hidden within traditional baseline comparisons.

Due to the limited target-class sizes under sever imbalance inherent to clinical data, computing AUGRC on individual test folds yields numerically unstable artifacts. To ensure evaluation stability, we assess the final aggregated ensemble via patient-level bootstrapping. Point estimates are reported with 95\% bootstrap confidence intervals (CIs) computed from $1,000$ patient-level resamples of the aggregated predictions. Pairwise statistical comparisons against the proposed $H_\tau$~+~CCR framework utilize a bootstrap sign test. To rigorously control the family-wise error rate (FWER) across the multiple baseline comparisons, we apply the Holm-Bonferroni step-down procedure independently for each disease cohort and evaluation metric. A performance difference is considered statistically significant only if its confidence interval strictly excludes zero and the comparison satisfies the adjusted Holm-Bonferroni threshold at family-wise $\alpha = 0.05$.

For the synthetic experiments, the analysis focuses on mapping performance across the grid of $\pi$ and $d$ conditions to establish the phase diagram. We report Sensitivity at a fixed 30\% rejection rate as the primary phase diagram metric, alongside the Sensitivity stability slope to delineate regimes of active minority deferral. Full rejection curve panels for all 56 synthetic conditions are provided in the Supplementary Material.

\section{Results}\label{sec3}
\subsection{Synthetic validation}\label{sec3.1}
To isolate and empirically validate the structural vulnerabilities of standard uncertainty quantification (Vulnerabilities 1 and 2), we first evaluated the frameworks across 56 highly controlled synthetic conditions. This allowed us to map the precise boundaries of algorithmic failure without the confounding effects of clinical biological heterogeneity. Empirical confirmation of Vulnerability 3 (structural uncertainty divergence) is provided in Section \ref{sec3.2}.
\subsubsection{Phase diagram of standard UQ failure}
\begin{figure}
    \centering
    \includegraphics[width=1\linewidth]{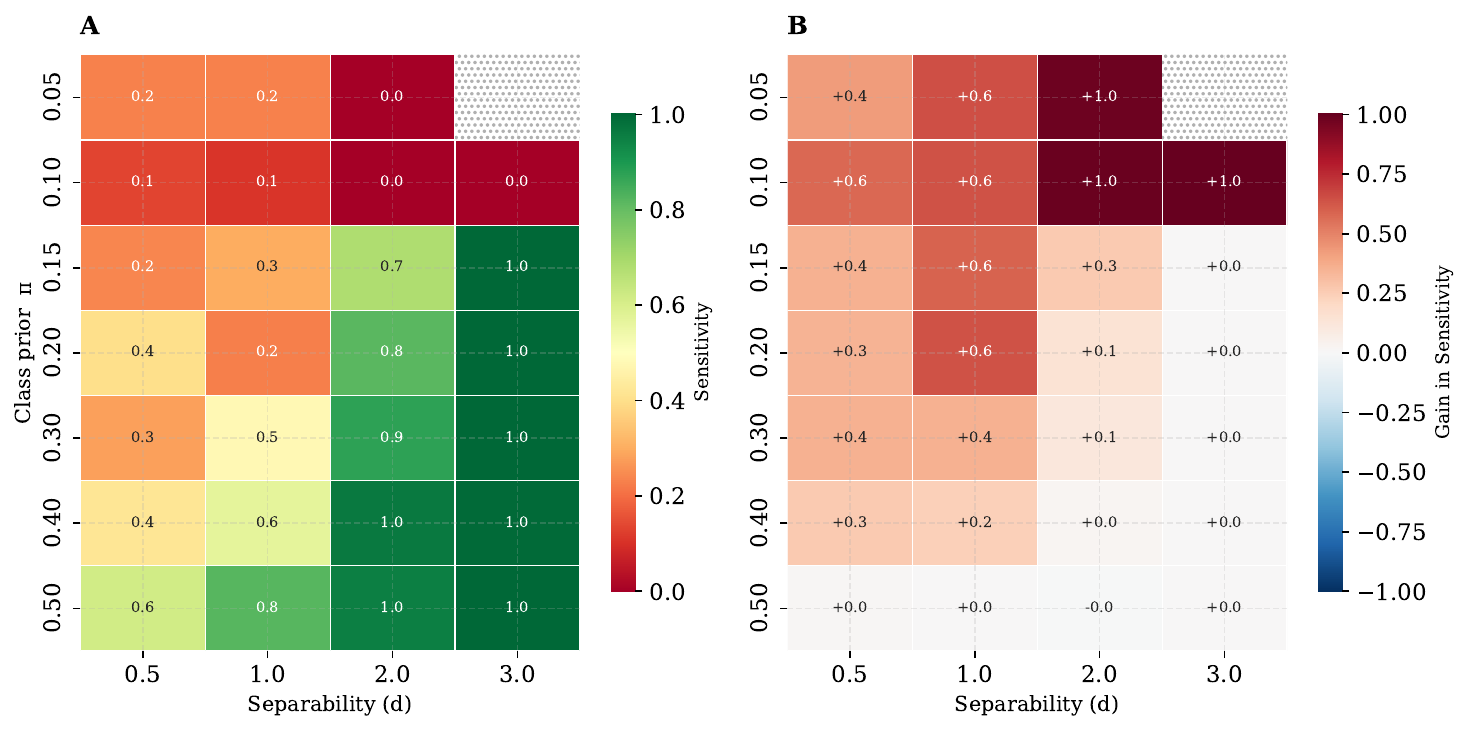}
    \caption{\textbf{Failure landscape of selective prediction under class imbalance.} Heatmaps illustrate minority class Sensitivity at a fixed 30\% global rejection rate across 56 synthetic conditions, parameterized by class prior ($\pi \in [0.05, 0.50]$, x-axis) and class separability ($d \in [0.5, 3.0]$, y-axis). \textbf{(A)} Standard global $H_{\text{Total}}$ exhibits a severe structural failure boundary. Under conditions of severe imbalance and overlapping distributions ($d \leq 1.0, \pi \leq 0.20$) the baseline disproportionately defers minority patients (Sensitivity $\approx 0$). \textbf{(B)} The proposed $H_\tau$~+~CCR dual-architecture successfully rescues minority coverage within this diagnostic failure zone. Crucially, the architectural correction scales proportionally with the degree of imbalance, smoothly converging to identical standard baseline behavior when the natural prior is perfectly balanced ($\pi=0.50$).}
    \label{fig:phase_diag}
\end{figure}

To establish the failure landscape, we tracked minority class Sensitivity (at a fixed 30\% rejection rate) across a grid of varying class priors ($\pi \in [0.05, 0.50]$) and class separability ($d \in [0.5, 3.0]$). The results for all conditions (with $N=1000$ samples) are shown in Figure \ref{fig:phase_diag}. Standard $H_{\text{Total}}$ maintains high Sensitivity only when classes are easily separable ($d \geq 2.0$) or near-balanced ($\pi \geq 0.40$). However, under severe imbalance ($\pi \leq 0.20$) and overlapping distributions ($d \leq 1.0$) standard UQ undergoes a complete collapse. At $\pi=0.10$ and $d=1.0$ baseline minority Sensitivity drops to $0.11$, indicating that standard entropy systematically excludes nearly all minority patients from the retained cohort. Under the most severe conditions ($\pi=0.05, d=2.0$), it yields a Sensitivity of exactly $0.00$. The proposed $H_\tau$~+~CCR framework successfully rescues minority coverage within this failure zone. By anchoring the entropy peak to $\tau$ (set to $\pi$) and ranking within trajectories, the framework recovers up to $+0.99$ Sensitivity at severe imbalance ($\pi=0.10, d=2.0$). Crucially, this architectural correction scales proportionally with the degree of imbalance, converging perfectly at parity: when the natural prior is balanced ($\pi=0.50$), the difference between the proposed framework and the baseline is exactly zero. Refer to Appendix \ref{secA3} for all conditions with $N=300$ samples. 

\subsubsection{Sensitivity stability scaling law}
\begin{figure}
    \centering
    \includegraphics[width=1\linewidth]{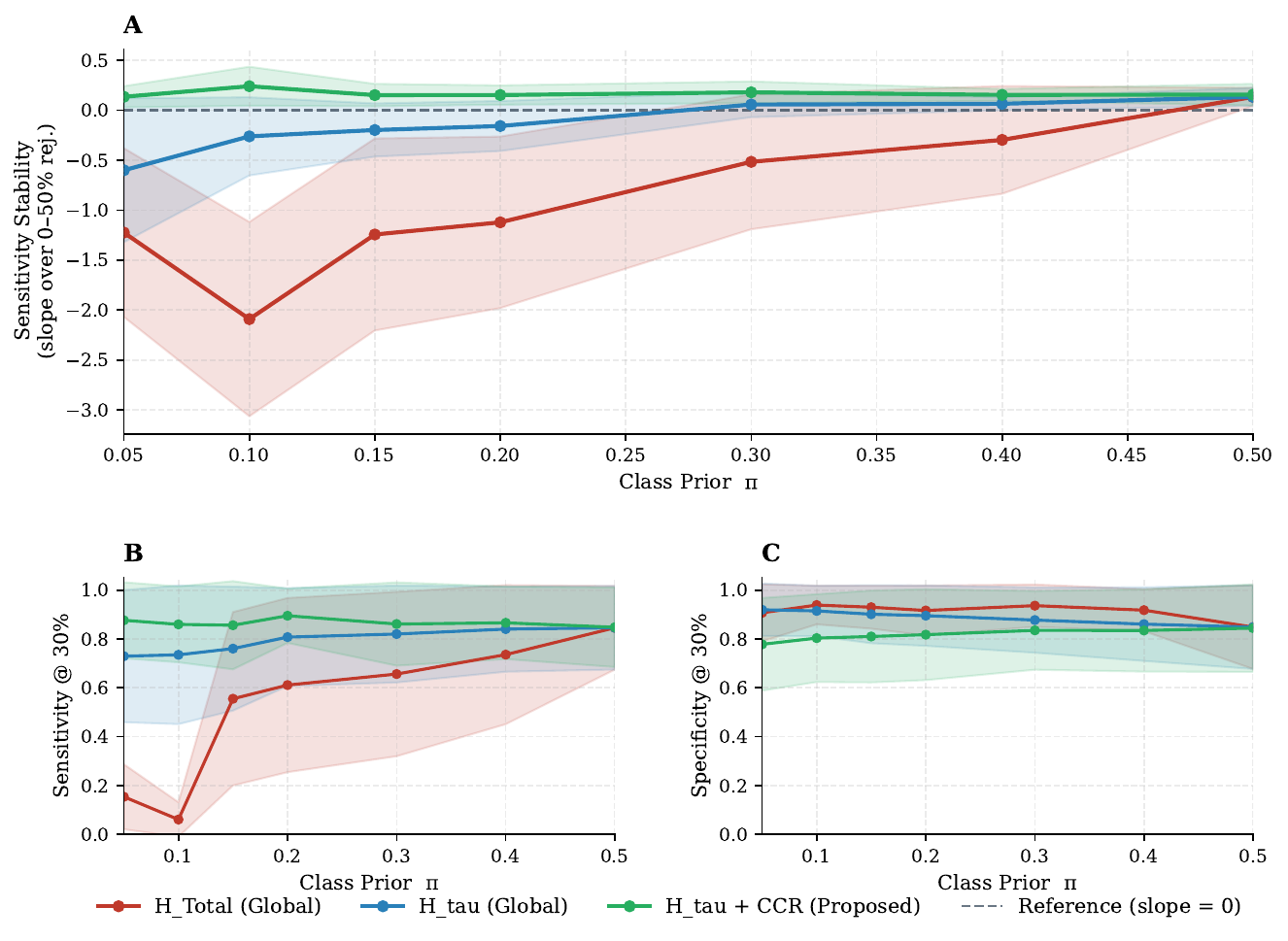}
    \caption{\textbf{Scaling law of minority Sensitivity under dynamic rejection across class priors.} Panels show the mean (solid lines) and standard deviation (shaded regions) aggregated across varying levels of intrinsic class separability ($d \in [0.5, 3.0]$) at a sample size of $n=1000$. \textbf{(A)} Sensitivity stability, defined as the linear slope of minority Sensitivity across the $0\%\text{--}50\%$ rejection range. Standard global $H_{\text{Total}}$ (red) exhibits a severe non-linear scaling law, plunging into severe minority exclusion as class imbalance increases. The proposed $H_\tau$~+~CCR dual-architecture (green) fully neutralizes this artifact, maintaining a safe trajectory near zero across all priors. 
    \textbf{(B)} Sensitivity evaluated at a fixed 30\% global rejection rate. Standard UQ systematically collapses under severe imbalance ($\pi \leq 0.20$), whereas the proposed framework guarantees robust minority retention. 
    \textbf{(C)} Specificity (majority class retention) evaluated at the exact same 30\% rejection rate. Under a fixed global rejection constraint, standard UQ maximizes Specificity by disproportionately concentrating deferrals within the minority class. By recovering minority Sensitivity (Panel B), the proposed framework mathematically redistributes these deferrals, resulting in a balanced and class-proportional rejection profile.}
    \label{fig:sensitivity_stability}
\end{figure}

To capture the dynamic failure of standard UQ, we evaluated the stability of Sensitivity across the continuous range of class priors (Figure \ref{fig:sensitivity_stability}A). Our analysis exposes a severe scaling law: systematic exclusion of minority patients by standard UQ scales inversely and non-linearly with class imbalance. While standard $H_{\text{Total}}$ maintains a neutral stability slope ($+0.13$) at perfect balance ($\pi=0.50$), it aggressively defers the minority class at extreme imbalance, driving the slope down to $-2.09$ at $\pi=0.10$. 

Evaluating the independent performance curves confirms that neither component alone can resolve this collapse. Applying only the $H_\tau$ signal globally successfully resolves the threshold-entropy mismatch (Vulnerability 1), sharply reducing the exclusion slope at $\pi=0.10$ from $-2.09$ to $-0.26$. However, because the spatial penalty (Vulnerability 2) remains unaddressed, the system still leaks minority patients. Implementing the complete dual-architecture ($H_\tau$~+~CCR) fully stabilizes the triage mechanism. The combined framework maintains a safe, positive stability slope (bounded between $+0.13$ and $+0.24$) across all tested priors, maintaining stable minority protection regardless of the underlying disease prevalence.

Evaluating the system at a fixed global rejection rate highlights the mechanistic trade-off inherent to selective classification (Figure \ref{fig:sensitivity_stability}C). Under severe imbalance, the standard baseline achieves a disproportionately high Specificity ($\approx 0.94$ at $\pi=0.10$) strictly by displacing its rejections onto the minority class. By correcting this structural bias to recover minority Sensitivity, the proposed $H_\tau$~+~CCR framework forces a mathematical reallocation of the deferrals. The resulting trade-off, which is a modest reduction in Specificity to secure a substantial recovery in minority Sensitivity, demonstrates a balanced distribution of uncertainty across both disease trajectories.

\subsection{Real-world validation}\label{sec3.2}
To validate these findings in clinical practice, we evaluated the framework across three real-world cohorts (MS, PD, AD) whose class priors ($\pi \in \{0.13, 0.39, 0.54\}$) strategically span the structural failure zones identified in the synthetic experiments.

\subsubsection{Empirical confirmation of uncertainty divergence}\label{sec3.2.1}
To validate the theoretical presence of Vulnerability 3 (structural uncertainty divergence), we analyzed the relationship between the predicted probabilities and the ensemble disagreement $\sigma^2$ (used as a proxy of epistemic uncertainty) across the clinical trajectories. 

\begin{figure}[htpb]
    \centering
    \includegraphics[width=1\linewidth]{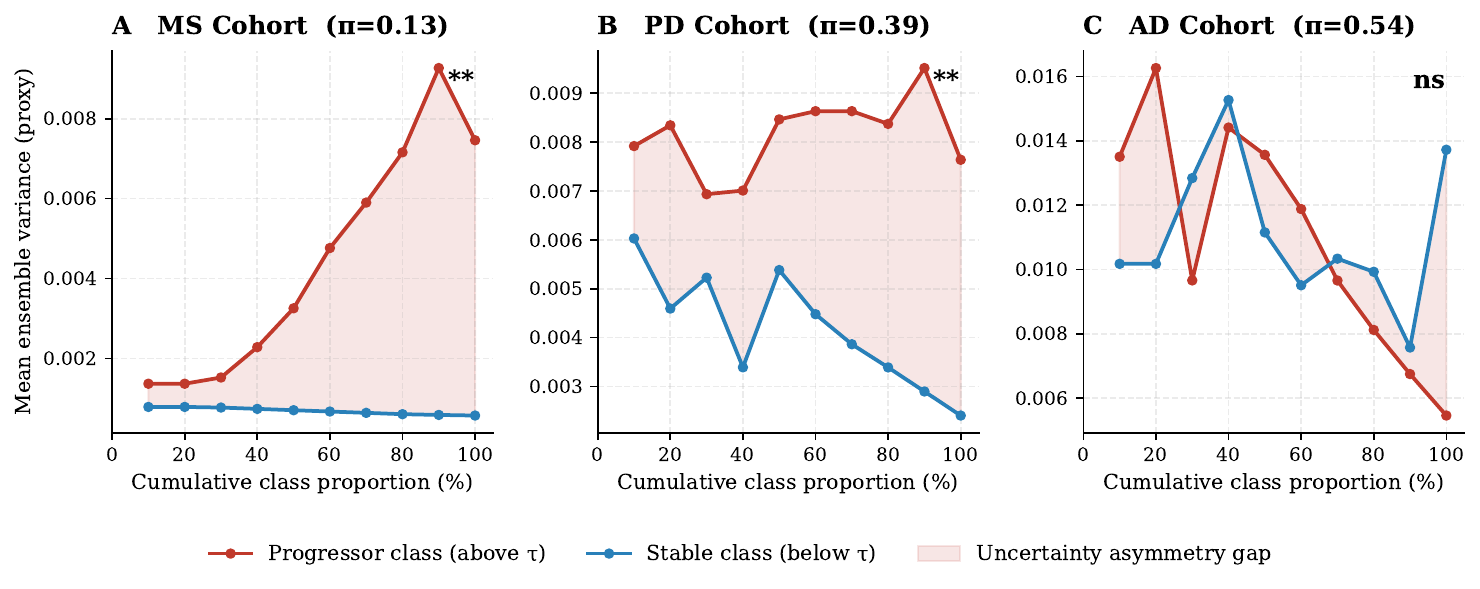}
    \caption{\textbf{Empirical confirmation of asymmetric uncertainty divergence.} Mean ensemble disagreement (used as a proxy for epistemic uncertainty) is evaluated as a function of the prediction's relative distance from the class-conditioned decision boundary, binned by class-specific distance deciles. Statistical significance is evaluated via a paired Wilcoxon signed-rank test across the distance deciles (*** $p < 0.001$, ** $p < 0.01$, * $p < 0.05$, ns: not significant). Because data is partitioned into deciles, the sample size per bin remains rigorously constant within each cohort. \textbf{(A)} Under severe class imbalance (MS cohort), a profound uncertainty asymmetry gap emerges (**). As minority progressor predictions (red) move away from the decision threshold, their variance proportionally increases, whereas majority stable predictions (blue) converge near zero ($n \approx 23$ minority / $48$ majority patients per decile bin). \textbf{(B)} In the moderately imbalanced PD cohort, minority variance remains structurally elevated across boundary distances (**; $n \approx 5$ minority / $9$ majority per bin). \textbf{(C)} In the near-balanced AD cohort, this asymmetry naturally dissolves (ns; $n \approx 8$ minority / $6$ majority per bin). This cross-cohort progression empirically confirms that uncertainty divergence is a mechanical artifact of class imbalance rather than intrinsic clinical noise.}
    \label{fig:asymmetry_test}
\end{figure}

Figure~\ref{fig:asymmetry_test} illustrates the highly asymmetric uncertainty profiles characteristic of imbalanced prognostic tasks. To formally test this divergence while accounting for the limited minority sample size, predictions were grouped into deciles based on their relative distance from the class-conditioned boundary (cumulative class proportions). For the densely sampled majority class (stable patients, $y=0$), the model demonstrates solid certainty: as predictions move further from the decision boundary, the ensemble disagreement smoothly converges to zero. In contrast, the sparse minority class lacks this representational density. Consequently, even when the model correctly assigns high-confidence probabilities to progressor patients, these predictions are accompanied by a structurally higher ensemble disagreement. A paired Wilcoxon signed-rank test across these distance deciles formally confirms this asymmetry. In the severely imbalanced MS cohort, the minority class exhibits significantly higher variance than the majority class at equivalent relative distances from the boundary ($p < 0.01$, with $n \approx 23$ minority and $n \approx 48$ majority patients per decile bin). This structural divergence persists in the moderately imbalanced PD cohort ($p < 0.01$) but naturally dissolves in the near-balanced AD cohort ($p = \text{n.s.}$).  This confirms that standard global ranking systems inherently penalize minority patients not due to inherent biological ambiguity (aleatoric noise), but primarily due to the epistemic sparsity of their cohort.

To isolate the effect of class frequency and confirm this divergence is not an artifact of varying clinical feature spaces, we replicated this exact analysis under tightly controlled synthetic conditions (Supplementary Figures \ref{fig:figC4_uncertainty_div_n300}, \ref{fig:figC4_uncertainty_div_n1000}). By fixing the underlying class separability at $d=1.0$ and exclusively varying the minority prevalence, we reproduced the identical structural pattern. At a high imbalance of $\pi = 0.15$, the uncertainty asymmetry gap between classes is highly significant ($p < 0.01$). As the prevalence is artificially increased to $\pi=0.30$, the gap narrows but remains significant ($p < 0.01$), before fully dissipating at parity ($\pi=0.50$, $p = \text{n.s.}$).

\subsubsection{Rejection curve analysis}
\begin{figure}
    \centering
    \includegraphics[width=1\linewidth]{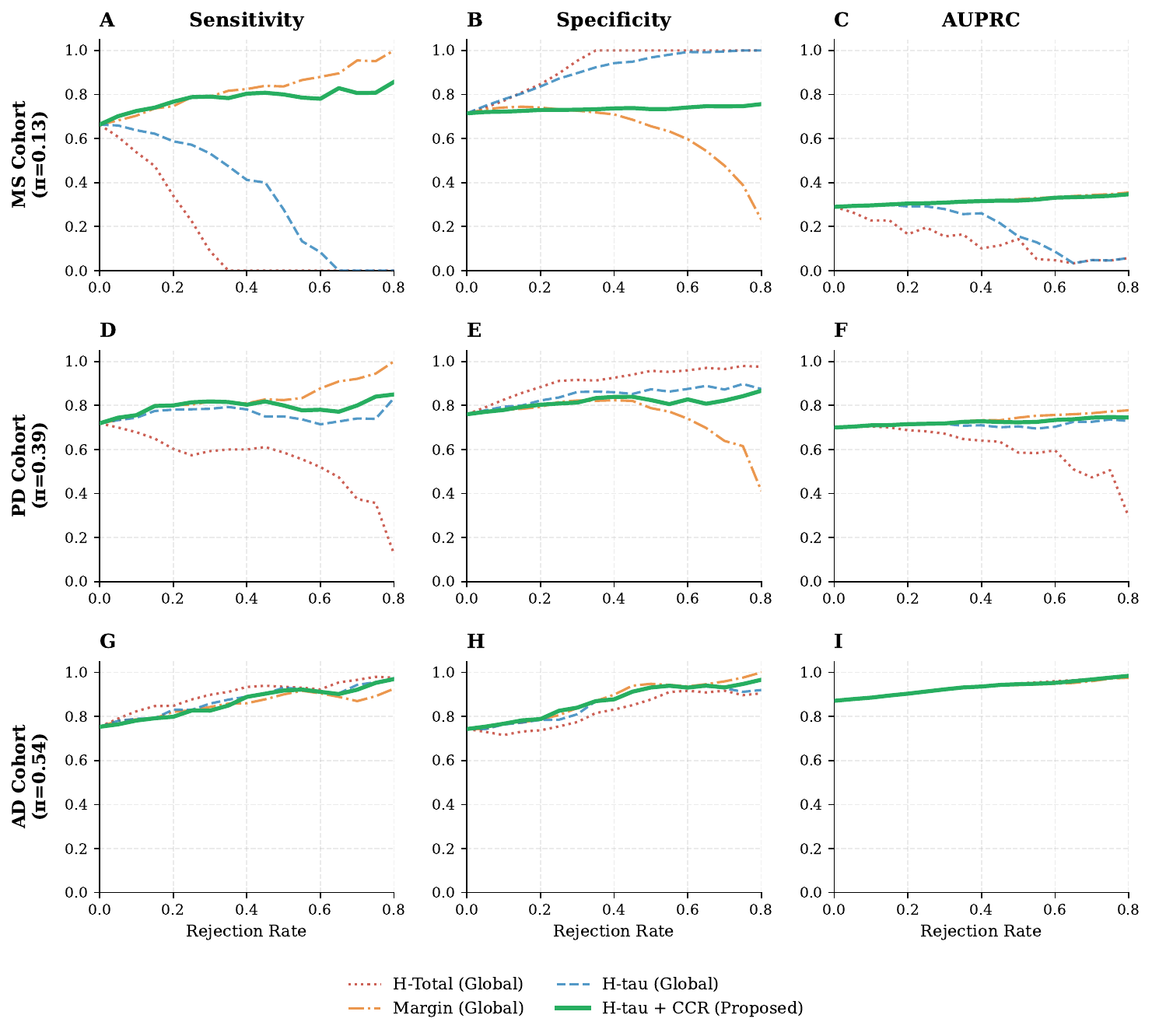}
    \caption{\textbf{Rejection trajectories across real-world clinical cohorts.} Performance is evaluated across the full continuum of rejection rates ($r \in [0, 0.80]$) for severe imbalance (MS, top row), moderate imbalance (PD, middle row), and near-balance (AD, bottom row). Under severe imbalance, standard global $H_{\text{Total}}$ (red dashed line) induces a complete Sensitivity collapse, systematically excluding minority progressor patients from the retained cohort. Conversely, the global Margin baseline (orange) empirically instantiates the spatial penalty: while it artificially protects minority Sensitivity, it systematically depletes majority class Specificity. The proposed $H_\tau$~+~CCR dual-architecture (green) breaks this Sensitivity-Specificity trade-off, strictly bounding risk for both clinical trajectories and safely improving overall clinical utility (AUPRC). In the near-balanced AD cohort, these structural vulnerabilities inherently dissolve, and all frameworks safely converge.}
    \label{fig:rejection_curves}
\end{figure}
To trace the dynamic degradation of target-class coverage, we evaluated class-conditioned performance trajectories across the full continuum of rejection rates ($r \in [0, 0.80]$) for the three clinical cohorts (Figure \ref{fig:rejection_curves}). These empirical curves visually confirm the structural vulnerabilities of standard uncertainty metrics under severe class imbalance. Under severe imbalance (MS cohort, $\pi=0.13$), $H_{\text{Total}}$ triggers a catastrophic failure in minority protection. As the rejection rate increases, $H_{\text{Total}}$ systematically excludes the progressor class, driving minority Sensitivity to zero by a 35\% rejection rate. While applying the global $H_\tau$ correction without class-conditioning delays this collapse by fixing the threshold-entropy mismatch, it ultimately falls victim to epistemic compression and spatial penalties, reaching zero Sensitivity by 60\% rejection. The Margin baseline instead empirically instantiates the spatial penalty: because the MS decision threshold ($\tau=0.13$) is tightly compressed against the majority class space, the Margin metric systematically rejects highly confident stable (majority) patients. While this artificially inflates minority Sensitivity, it triggers a collapse in majority Specificity, which plunges from 0.70 down to nearly 0.20 at high rejection rates. This Sensitivity-Specificity trade-off, where an algorithm can only protect the minority by systematically reducing majority class coverage, renders global Margin structurally and practically unviable for these conditions. 
Combining CCR with a threshold-aware metric, such as our proposed $H_\tau$ (or the equivalent Margin), is the only framework that breaks this dynamic. By generating a centered uncertainty signal and enforcing class-conditioned triage queues, the framework smoothly and simultaneously improves both target-class Sensitivity and stable-class Specificity. 

Empirical verification of the internal queue dynamics confirmed this continuous, symmetric error-filtering mechanism across the full continuum of rejection rates. For example, at a standard 30\% global rejection rate in the MS cohort, the framework deferred false negatives from the $\mathcal{D}_0$ queue at more than twice the rate of true negatives (58.1\% vs. 27.9\%). Simultaneously, it actively reduced the number of false positives in the $\mathcal{D}_1$ queue at a substantially higher rate than confident true positives (33.5\% vs. 19.7\%). These targeted filtering trends hold across all evaluated cohorts (Supplementary Figure \ref{fig:figD1_reject_dynamics}). This mechanism drives overall predictive utility (AUPRC) upward as uncertain predictions are safely deferred. These structural trends persist under moderate imbalance (PD cohort, $\pi=0.39$), where standard UQ continually erodes minority Sensitivity while the proposed framework maintains stable, bounded protection for both classes. Finally, on the near-balanced AD cohort ($\pi=0.54$), the structural vulnerabilities dissolve, and all uncertainty frameworks smoothly converge to nearly identical, safe trajectories.

\subsubsection{Class-conditioned safety evaluation}
Evaluation in real-world clinical cohorts demonstrates that standard global uncertainty quantification fails under class imbalance, a vulnerability masked by standard global metrics (Table~\ref{tab_risk_results}). To rigorously map the statistical stability of the generalized risk across patient trajectories, we performed post-prediction non-parametric bootstrapping ($B=1000$ iterations) and calculated the class-conditioned Area Under the Generalized Risk-Coverage Curve (ccAUGRC). Under severe imbalance (MS) and moderate imbalance (PD), standard $H_{\text{Total}}$ registers a biased statistical risk for the progressor class. While the Margin baseline appears to improve progressor risk, it simply displaces the error, resulting in statistically worse stable class ccAUGRC. Applying CCR with a threshold-aware metric (instantiated here by $H_\tau$~+~CCR) is the only framework that successfully bounds risk for both clinical trajectories simultaneously. The Random~+~CCR control isolates the contribution of the class-conditioned rejection mechanism from the uncertainty signal. At MS, Random~+~CCR achieves a progressor ccAUGRC of 0.1605 [0.1084, 0.2123], significantly better than $H_{\text{Total}}$ ($p<0.001$) and global $H_\tau$ ($p=0.004$), but significantly worse than $H_\tau$~+~CCR ($p<0.001$), a performance gap that holds even after Holm-Bonferroni correction.
Furthermore, the $H_\text{Total}$~+~CCR ablation demonstrates that enforcing class-preservation on a misaligned global uncertainty signal is insufficient: it yields a progressor ccAUGRC of 0.1467, remaining significantly worse than the fully proposed framework ($p=0.008$ after FWER adjustment). 
This decomposition quantifies the independent contribution of each component: the class-preservation mechanism alone accounts for the majority of the improvement over standard methods, while the threshold-aware $H_\tau$ signal provides additional gain by correctly ranking the most threshold-ambiguous patients within each trajectory. 
To ensure our partial-area formulation does not bias this comparative evaluation, all metrics were recalculated over a strict common support (Supplementary Table \ref{tab_risk_results_cs}), fully confirming this relative performance ordering.

The near-balanced AD cohort ($\pi=0.54$) validates the framework's behavior outside of the failure zone. In near-balance, $H_{\text{Total}}$ achieves a lower progressor ccAUGRC than $H_\tau$~+~CCR (0.0567 vs 0.0767, $p<0.001$ adjusted), while simultaneously showing significantly higher stable ccAUGRC (0.1008 vs 0.0748, $p<0.001$ adjusted). This reflects the exact inter-class substitution artifact identified in the Methods: at $\tau= \pi = 0.54$, where progressors are the  majority class, $H_{\text{Total}}$ preferentially retains progressors at the cost of stable patients rather than excluding them as it does at lower $\tau$. As explicitly quantified by the class disparity column in Table \ref{tab_risk_results}, standard global entropy yields a highly skewed risk profile (disparity = 0.0441), even while its global AUGRC remains superficially competitive.
The proposed $H_\tau$~+~CCR framework prevents this unchecked substitution and reduces this disparity to near-zero (0.0019), preserving class-proportional risk profiles across both classes. In this near-balance scenario, the proposed architecture acts as a robust sanity check, safely matching the performance of standard methods without introducing detrimental artifacts, while the Random~+~CCR baseline naturally degrades to uninformative deferral.

\begin{table*}[t]
\caption{\textbf{Generalized risk-coverage evaluation across clinical cohorts.} Performance is measured via the class-conditioned Area Under the Generalized Risk-Coverage Curve (ccAUGRC) and Global AUGRC. The class disparity column ($|\text{Progressor ccAUGRC} - \text{Stable ccAUGRC}|$) explicitly quantifies between-class equity. Values are point estimates with 95\% Confidence Intervals derived from 1000 post-prediction bootstrap iterations. Significance testing compares each baseline to the proposed reference framework ($H_\tau$~+~CCR), with $p$-values adjusted for multiple comparisons using the Holm-Bonferroni method. Applying CCR with a threshold-aware metric (instantiated here by $H_\tau$~+~CCR) is the only evaluated method to safely bound risk for both clinical trajectories under severe imbalance while minimizing class disparity. Lower values indicate better safety/lower risk.}\label{tab_risk_results}
\centering
\resizebox{\textwidth}{!}{
\begin{tabular}{l l l l l l}
\toprule
Dataset & Uncertainty Framework & Progressor ccAUGRC (95\% CI) & Stable ccAUGRC (95\% CI) & Class Disparity & Global AUGRC (95\% CI) \\
\midrule
\multirow{6}{*}{\textbf{MS} ($\tau=\pi=0.13$)} & $H_{\text{Total}}$ (Global) & 0.2756 [0.2080, 0.3357]$^{\uparrow\textcolor{red}{***}}$ & 0.0413 [0.0317, 0.0518]$^{\downarrow\textcolor{red}{***}}$ & 0.2343 & 0.0679 [0.0556, 0.0813] \\
  & Margin (Global) & 0.0931 [0.0562, 0.1378]$^{\downarrow\textcolor{red}{**}}$ & 0.1529 [0.1313, 0.1750]$^{\uparrow\textcolor{red}{***}}$ & 0.0598 & 0.1431 [0.1235, 0.1625] \\
  & $H_\tau$ (Global) & 0.2178 [0.1535, 0.2762]$^{\uparrow\textcolor{red}{***}}$ & 0.0561 [0.0449, 0.0680]$^{\downarrow\textcolor{red}{***}}$ & 0.1617 & 0.0734 [0.0605, 0.0867] \\
  & Random~+~CCR & 0.1605 [0.1084, 0.2123]$^{\uparrow\textcolor{red}{***}}$ & 0.1374 [0.1201, 0.1538]$^{\uparrow\textcolor{red}{**}}$ & 0.0231 & 0.1404 [0.1231, 0.1577] \\
  & $H_{\text{Total}}$~+~CCR & 0.1467 [0.0941, 0.2013]$^{\uparrow\textcolor{red}{**}}$ & 0.1395 [0.1230, 0.1562]$^{\uparrow\textcolor{red}{*}}$ & 0.0072 & 0.1404 [0.1235, 0.1577] \\
  & \textbf{$H_\tau$~+~CCR} & \textbf{0.1115 [0.0682, 0.1564]} & \textbf{0.1301 [0.1135, 0.1466]} & \textbf{0.0186} & \textbf{0.1275 [0.1119, 0.1442]} \\
\midrule
\multirow{6}{*}{\textbf{PD} ($\tau=\pi=0.39$)} & $H_{\text{Total}}$ (Global) & 0.1891 [0.1288, 0.2469]$^{\uparrow\textcolor{red}{***}}$ & 0.0431 [0.0218, 0.0660]$^{\downarrow\textcolor{red}{***}}$ & 0.1460 & 0.0918 [0.0646, 0.1191] \\
  & Margin (Global) & 0.0840 [0.0454, 0.1275]$^{\downarrow\textcolor{red}{**}}$ & 0.1066 [0.0681, 0.1468]$^{\uparrow\text{ns}}$ & 0.0226 & 0.0973 [0.0688, 0.1255] \\
  & $H_\tau$ (Global) & 0.1160 [0.0676, 0.1651]$^{\uparrow*}$ & 0.0776 [0.0469, 0.1118]$^{\downarrow*}$ & 0.0384 & 0.0926 [0.0649, 0.1188] \\
  & Random~+~CCR & 0.1351 [0.0890, 0.1851]$^{\uparrow*}$ & 0.1147 [0.0798, 0.1505]$^{\uparrow\textcolor{red}{*}}$ & 0.0204 & 0.1228 [0.0940, 0.1541] \\
  & $H_{\text{Total}}$~+~CCR & 0.1052 [0.0600, 0.1519]$^{\uparrow\text{ns}}$ & 0.0950 [0.0612, 0.1306]$^{\uparrow\text{ns}}$ & 0.0102 & 0.0987 [0.0720, 0.1267] \\
  & \textbf{$H_\tau$~+~CCR} & \textbf{0.1031 [0.0577, 0.1494]} & \textbf{0.0912 [0.0564, 0.1259]} & \textbf{0.0119} & \textbf{0.0958 [0.0681, 0.1227]} \\
\midrule
\multirow{6}{*}{\textbf{AD} ($\tau=\pi=0.54$)} & $H_{\text{Total}}$ (Global) & 0.0567 [0.0352, 0.0834]$^{\downarrow\textcolor{red}{***}}$ & 0.1008 [0.0676, 0.1382]$^{\uparrow\textcolor{red}{***}}$ & 0.0441 & 0.0756 [0.0570, 0.0966] \\
  & Margin (Global) & 0.0800 [0.0525, 0.1115]$^{\uparrow\text{ns}}$ & 0.0733 [0.0459, 0.1057]$^{\downarrow\text{ns}}$ & 0.0067 & 0.0763 [0.0572, 0.0966] \\
  & $H_\tau$ (Global) & 0.0708 [0.0463, 0.1001]$^{\downarrow\text{ns}}$ & 0.0818 [0.0520, 0.1162]$^{\uparrow\text{ns}}$ & 0.0110 & 0.0755 [0.0567, 0.0964] \\
  & Random~+~CCR & 0.1180 [0.0852, 0.1540]$^{\uparrow\textcolor{red}{***}}$ & 0.1228 [0.0859, 0.1626]$^{\uparrow\textcolor{red}{***}}$ & 0.0048 & 0.1203 [0.0940, 0.1479] \\
  & $H_{\text{Total}}$~+~CCR & 0.0773 [0.0521, 0.1068]$^{\uparrow\text{ns}}$ & 0.0755 [0.0473, 0.1071]$^{\uparrow\text{ns}}$ & 0.0018 & 0.0764 [0.0573, 0.0973] \\
  & \textbf{$H_\tau$~+~CCR} & \textbf{0.0767 [0.0515, 0.1067]} & \textbf{0.0748 [0.0465, 0.1061]} & \textbf{0.0019} & \textbf{0.0758 [0.0570, 0.0966]} \\
\bottomrule
\end{tabular}
}
\vspace{1ex}
{\raggedright \footnotesize
Significance vs. Proposed ($H_\tau$~+~CCR): *** $p<0.001$, ** $p<0.01$, * $p<0.05$, ns: not significant. \\
Arrows indicate if the baseline is structurally worse ($\uparrow$, higher risk) or better ($\downarrow$, lower risk). \\
Red markers ($\textcolor{red}{*}$) indicate that statistical significance is maintained after Holm-Bonferroni correction ($\alpha=0.05$). \par}
\end{table*}

\newpage
\section{Discussion}\label{sec12}

Our findings demonstrate that under severe class imbalance, standard uncertainty quantification (UQ) fundamentally ceases to measure clinically aligned ambiguity, instead functioning as a proxy for class membership. Evaluation in synthetic and real-world neurodegenerative cohorts confirms that global metrics, such as Shannon entropy and classification margin, structurally penalize minority patients. This penalization is driven by three compounding mathematical failure modes: threshold-entropy mismatch, asymmetric spatial penalties, and structural uncertainty divergence. 

Clinically, this measurement failure has severe implications. In prognostic settings, the minority class often represents the most severe and clinically urgent trajectories, such as rapid progressors in multiple sclerosis and Parkinson's Disease. 

Because minority patients are typically represented by sparser feature distributions, they exhibit structurally higher ensemble disagreement. By conflating this dataset sparsity with decision-level ambiguity, standard global UQ systems systematically defer high-risk patients, actively excluding them from the automated pipeline and creating an untenable algorithmic trade-off. 

The empirical phase diagram defines clear boundaries under which threshold-aware, class-conditioned frameworks (such as $H_\tau$~+~CCR) are strictly necessary, namely when class prior and class separability jointly fall below $\pi \approx 0.20$ and $d \approx 1.0$.
In this region, standard metrics lead to Sensitivity collapse under severe imbalance and to progressive minority erosion at moderate imbalance. Both components of the dual-architecture are therefore required to recover system performance. The proposed threshold-aware ambiguity score ($H_\tau$) corrects the signal layer by realigning maximum uncertainty with the natural class prior, while Class-Conditioned Rejection (CCR) corrects the triage layer by ranking patients within their predicted trajectories. In doing so, CCR neutralizes the structurally elevated absolute uncertainty floor associated with the minority class. 

Crucially, the ablation study using the Random~+~CCR control isolates the contribution of the class-conditioned rejection mechanism. The results show that class-preservation alone substantially mitigates minority exclusion, while the $H_\tau$ signal provides an additional benefit by correctly ordering ambiguous cases within those protected queues.

Conversely, in near-balanced clinical tasks, such as the Alzheimer's Disease cohort ($\pi = 0.54$), these structural vulnerabilities naturally attenuate. Under these balanced conditions, the proposed framework  converges to standard baseline behavior, indicating that the architectural correction scales proportionally with the degree of imbalance  and introduces no measurable penalty when the classes are approximately balanced. Notably, even in this near-balanced setting, $H_{\text{Total}}$ exhibits a residual substitution artifact in the opposite direction, preferentially protecting the now-larger progressor class at the expense of stable patients. This suggests that inter-class substitution is a general property of global aggregation, rather than a phenomenon specific to severe imbalance.

These findings extend recent theoretical work questioning the appropriateness of conditional entropy and mutual information as uncertainty measures under distribution shift \cite{wimmer2023quantifying, jimenez2026position}. While prior work has largely focused on probabilistic recalibration or multimodal data acquisition to improve predictive confidence, these results show that standard selective failure is distinctly geometric. This vulnerability persists even after rigorous Platt scaling because it arises fundamentally from the structural misalignment between entropy, the shifted decision boundary, and the asymmetric probability space. 

A core consequence of these vulnerabilities is that standard evaluation paradigms, such as the Area Under the Generalized Risk-Coverage Curve (AUGRC), may fail to detect automated minority exclusion. Because global multi-threshold metrics aggregate errors across the full population, a selective prediction system can mathematically improve its overall safety score by shifting  residual risk onto the harder-to-predict minority class. To address this inter-class substitution artifact, we introduced class-conditioned AUGRC (ccAUGRC). By quantifying the generalized risk-coverage trade-off separately for each clinical trajectory, ccAUGRC prevents risk-masking and makes subgroup-specific failures directly observable. This addresses a critical gap identified by recent literature \cite{pugnana2023auc, traub2024overcoming}, which successfully improved multi-threshold metrics but had yet to decompose risk by class subpopulation. Integrating such class-stratified metrics into reproducible machine learning pipelines is crucial to ensure that automated classification mechanisms maintain strictly bounded risk across all subgroups of patients, rather than concealing algorithmic bias behind improved overall safety.

Operationalizing the CCR framework within clinical workflows requires moving from absolute uncertainty thresholding to queue-based triage. Rather than attempting to mathematically optimize a global cost function, which requires assigning precise relative weights to false negatives, false positives, and clinician workload, CCR operationalizes deferral strictly as a fixed resource constraint. In practice, the hospital system first defines the decision threshold ($\tau$) according to the observed disease prevalence ($\pi$) in its target population. Incoming patient predictions are then assigned to their respective diagnostic trajectories and ranked by their $H_\tau$ scores. Administrators can subsequently define distinct, clinically appropriate rejection rates ($r$) for each queue based on available expert capacity. This ensures that manual review resources are allocated proportionally across the cohort, ensuring that the most ambiguous cases from every diagnostic trajectory receive expert review, rather than allowing one class to monopolize the triage bandwidth or erroneously automating boundary cases. Crucially, however, CCR bases triage decisions exclusively on predicted trajectories rather than true clinical states. Consequently, a true progressor confidently misclassified by the base classifier as stable (a false negative with $p \ll \tau$) is assigned to the predicted-stable queue and does not benefit from the protective balancing mechanisms applied to the progressor queue. CCR is not designed to correct confident classifier errors; rather, its specific, narrower purpose is to prevent the selective prediction mechanism from systematically starving the minority class of manual review capacity due to the structural geometry of the uncertainty score.

Positioning our framework relative to the conformal prediction literature clarifies the practical trade-offs of our approach. Class-conditional conformal prediction (CCCP) and distribution-free risk control provide rigorous, finite-sample guarantees on error rates by dynamically adapting the size of prediction sets. While theoretically optimal for bounding risk, this adaptive approach often yields variable rejection volumes that are difficult to align with fixed hospital resources (e.g., a specialist can only review a strict quota of 10 cases per day). Furthermore, under severe class imbalance, CCCP often satisfies its mathematical guarantees by generating uninformative prediction sets for the minority class (e.g., empty sets or multi-label predictions), which complicates binary clinical triage. Conversely, CCR sacrifices formal distribution-free coverage bounds in favor of strict operational constraints. By combining a fixed quota system with threshold-aware uncertainty ranking, CCR allows clinics to set a predictable budget for manual chart reviews (e.g., deferring exactly 30\% of cases) while structurally ensuring that this fixed bandwidth is distributed proportionally across predicted clinical trajectories, thereby preventing minority exclusion.

Although the present evaluation sets $\tau$ equal to the observed class prior $\pi$, in line with standard clinical practice, the proposed framework does not depend on this equivalence. By construction, $H_\tau$ reaches its maximum at $p = \tau$ for any $\tau \in (0,1)$, regardless of how $\tau$ is selected. In clinical settings, where false negatives carry substantially greater cost than false positives (such as multiple sclerosis screening, where missing a rapid progressor may be more harmful than unnecessarily flagging a stable patient), clinicians may deliberately set $\tau$ below $\pi$ to prioritize Sensitivity. The piecewise rescaling and class-conditioned rejection mechanism remain valid under this shift: $H_\tau$ continues to anchor maximum uncertainty to the selected operating point, while CCR preserves proportional representation around the resulting decision structure. The framework therefore generalizes to cost-sensitive threshold selection without requiring modification. However, when $\tau$ is shifted substantially away from $\pi$, the queue sizes preserved by CCR still reflect the empirical dataset distribution rather than the shifted clinical prior. Practitioners should account for this distinction when interpreting the resulting triage proportions.

Several limitations should be acknowledged. First, the framework is currently formulated for binary prognostic tasks; extension to multiclass settings will require rethinking both the entropy rescaling procedure, including which $\tau$ value anchors $H_{\tau}$ in the presence of multiple boundaries, and the CCR partitioning scheme. 
Furthermore, while the intra-class ordinal ranking of CCR makes it highly robust to monotonic miscalibration, severe violations of the logistic assumption during post-hoc calibration (e.g., Platt scaling) could distort absolute probabilities enough to shift boundary patients across the decision threshold $\tau$, thereby placing them into the incorrect triage queue. Additionally, the decision threshold $\tau$ is treated as known and fixed, and is set equal to the observed class prior. In deployment, however, $\tau$ may vary as disease prevalence changes or as patient populations differ across clinical sites. Regarding data availability, the Italian MS and PD cohorts cannot be publicly shared because of patient privacy regulations; however, we release the synthetic data and use the publicly available ADNI cohort in our analysis. Computationally, the ensemble pipeline is more demanding than single-model approaches, and the behavior of $H_\tau$~+~CCR in single-model or deep learning settings remains to be investigated. Finally, because the bootstrap procedure was applied post-prediction to the aggregated ensemble outputs, the resulting confidence intervals capture test-set sample variability but do not account for the variance of the model training process itself.

\section{Conclusion}\label{sec13}
The fundamental premise of selective prediction is that algorithms should defer to human experts when their predictions are uncertain. This study shows, however, that under severe clinical class imbalance, this safety mechanism can be compromised by the way uncertainty is operationalized. In these settings, standard uncertainty quantification may fail to capture clinically aligned ambiguity and instead reflect dataset sparsity.

The primary takeaway of this work is that reliance on standard, global uncertainty metrics can create a misleading illusion of safety. Because probabilistic models are structurally less confident for underrepresented classes, minority patients may receive systematically higher uncertainty scores that reflect data sparsity rather than triage-relevant ambiguity. Standard triage systems can therefore defer these patients at disproportionately high rates, while global evaluation metrics fail to detect the resulting bias, instead reporting improved overall performance simply by routing the most clinically urgent cases to the manual review queue.

By showing that combining a threshold-aware metric ($H_\tau$) with CCR mitigates this failure mode, our results demonstrate that this algorithmic bias is not an unavoidable consequence of imbalanced data, but a correctable limitation in how uncertainty is measured, ranked, and operationalized. By ensuring that rejection is applied proportionally across predicted trajectories (up to finite-sample rounding errors), and by realigning the uncertainty signal to aggressively filter false negatives near the decision boundary, clinical pipelines can safely defer ambiguous cases without systematically excluding the minority class from the automated pipeline.

These findings also highlight the need to distinguish between AI systems designed for biomarker discovery and those intended for clinical deployment. In exploratory discovery, high predictive variance might provide useful evidence of biological heterogeneity. In automated clinical triage, however, uncorrected variance can become a mechanism that deprives minority patients of the benefits of automation. Future clinical machine learning pipelines should therefore move beyond global risk-coverage evaluation and incorporate class-stratified metrics such as ccAUGRC. Algorithmic safety cannot be achieved by simply shifting the entire minority class to human experts; it requires risk to be directly measured and controlled within each patient trajectory, rather than inferred from the population as a whole.

\backmatter

\section{Declarations}
\subsection{Funding}
This work was partially funded under the National Plan for Complementary Investments to the NRRP, project “D34H—Digital Driven Diagnostics, prognostics and therapeutics for sustainable Health care” (project code: PNC0000001), Spoke 2: “Multilayer platform to support the generation of the Patients’ Digital Twin”, CUP: B53C22006170001, funded by the Italian Ministry of University and Research.

\subsection*{Conflict of interest}
The authors declare that they have no competing interests.
\subsection*{Ethics approval and consent to participate}
Not applicable.
\subsection*{Consent for publication}
Not applicable.
\subsection*{Data availability}
The data that support the findings of this study are available from the Alzheimer's Disease Neuroimaging Initiative (ADNI) but restrictions apply to the availability of these data, which were used under license for the current study, and so are not publicly available.

The remaining datasets analysed during the current study are not publicly available due to patient privacy and institutional clinical regulations, but are available from the corresponding author on reasonable request. These private datasets, including their baseline characteristics and validation protocols, have been previously described in detail in \cite{poretto2025machine} and \cite{Endrizzi2026Explainable}.

\subsection*{Code availability}
The code generated and analysed during the current study is available in a GitHub repository, \url{https://github.com/endrwalter/uncertainty_ml}.

\subsection*{Acknowledgements}
Data collection and sharing for the Alzheimer's Disease Neuroimaging Initiative (ADNI) is funded by the National Institute on Aging (National Institutes of Health Grant U19 AG024904). The grantee organization is the Northern California Institute for Research and Education. In the past, ADNI has also received funding from the National Institute of Biomedical Imaging and Bioengineering, the Canadian Institutes of Health Research, and private sector contributions through the Foundation for the National Institutes of Health (FNIH) including generous contributions from the following: AbbVie, Alzheimer's Association; Alzheimer's Drug Discovery Foundation; Araclon Biotech; BioClinica, Inc.; Biogen; Bristol-Myers Squibb Company; CereSpir, Inc.; Cogstate; Eisai Inc.; Elan Pharmaceuticals, Inc.; Eli Lilly and Company; EuroImmun; F. Hoffmann-La Roche Ltd and its affiliated company Genentech, Inc.; Fujirebio; GE Healthcare; IXICO Ltd.; Janssen Alzheimer Immunotherapy Research \& Development, LLC.; Johnson \& Johnson Pharmaceutical Research \& Development LLC.; Lumosity; Lundbeck; Merck \& Co., Inc.; Meso Scale Diagnostics, LLC.; NeuroRx Research; Neurotrack Technologies; Novartis Pharmaceuticals Corporation; Pfizer Inc.; Piramal Imaging; Servier; Takeda Pharmaceutical Company; and Transition Therapeutics.

\subsection*{Author contribution}
WE conceptualized the study, developed the methodology and software, and wrote the original draft of the manuscript. FR, RM, SB, and MM contributed to data curation, methodological development, and the interpretation of results, with FR additionally handling data visualization. FR and VO critically reviewed and finalised the manuscript. GJ and VO provided senior supervision and project administration. All authors read and approved the final manuscript.
\noindent

\newpage

\begin{appendices}
\section{ADNI Cohort Preprocessing}\label{secA1}
\subsection{Data Source and Ethics Statement}
Data used in the preparation of this article were obtained from the Alzheimer’s Disease Neuroimaging Initiative (ADNI) database (adni.loni.usc.edu). The ADNI was launched in 2003 as a public-private partnership, led by Principal Investigator Michael W. Weiner, MD. The primary goal of ADNI has been to test whether serial magnetic resonance imaging (MRI), positron emission tomography (PET), other biological markers, and clinical and neuropsychological assessment can be combined to measure the progression of mild cognitive impairment (MCI) and early Alzheimer’s disease (AD). For up-to-date information, see www.adni-info.org.  

\subsection{Cohort Selection Criteria}
To evaluate the proposed uncertainty framework on a near-balanced prognostic task ($\pi=0.54$), we formulated a 36-month conversion prediction task. We selected subjects who had a baseline diagnosis of Mild Cognitive Impairment (MCI). The longitudinal diagnostic trajectories of these subjects were tracked to establish the binary outcome:
\begin{itemize}
    \item Class 1 (Progressors / Converters): Subjects who converted to a diagnosis of AD within 36 months of their baseline visit ($N=170$).
    \item Class 0 (Stable / Non-converters): Subjects who maintained an MCI diagnosis for a minimum of 36 months, without reverting to cognitively normal or converting to AD ($N=144$).
\end{itemize}
Subjects who dropped out of the ADNI study or were lost to follow-up before the 36-month mark without a prior conversion to AD were strictly excluded from the analytic cohort, as their 36-month prognostic ground truth could not be definitively ascertained. This curation resulted in the final analytic cohort of $N=314$ patients. Descriptive statistics of the baseline characteristics are provided in Table~\ref{tab_adni_baseline}.
 
\subsection{Feature Engineering and Preprocessing}
To simulate a real-world clinical forecasting scenario, only data acquired at the baseline visit was utilized to predict the 36-month outcome. The predictive feature space encompassed demographic variables (e.g., age, sex, years of education) and established clinical/cognitive assessments (e.g., Mini-Mental State Examination [MMSE], Clinical Dementia Rating Sum of Boxes [CDR-SB], and the Alzheimer's Disease Assessment Scale-Cognitive Subscale [ADAS-Cog]).
 
Before model training, the data set was subjected to the following preprocessing pipeline. 
\begin{itemize}
    \item Missing Data Filtration: Features with a missing data proportion exceeding 50\% across the analytic cohort were excluded from the analysis.
    \item Imputation: To preserve sample size, missing baseline values for the retained features were imputed. Continuous variables were imputed using the median of the training distribution, while categorical variables were imputed using the mode (most frequent category). Imputation statistics were calculated strictly on the training folds during cross-validation to prevent data leakage.
    \item Scaling: All continuous numerical features were standardized by removing the mean and scaling to unit variance ($z$-score normalization), ensuring uniform weight distribution during model optimization.
\end{itemize}

\begin{table*}[t]
\caption{\textbf{Baseline Characteristics of the ADNI Analytic Cohort.} Categorical variables are presented as count (\%). Continuous variables are presented as median (Q1, Q3). P-values are calculated using the Mann-Whitney U test for continuous variables and the Pearson's chi-square test for categorical variables. Abbreviations: CDR-SB = Clinical Dementia Rating Sum of Boxes, FAQ = Functional Activities Questionnaire. Gender encoding in the dataset: 0.0 = Male, 1.0 = Female.}\label{tab_adni_baseline}
\centering
\begin{tabular}{l c c c c}
\toprule
Feature & Stable MCI & MCI-to-AD & Total Cohort & p-value \\
 & (N=144) & (N=170) & (N=314) & \\
\midrule
Age (years)             & 74.4 (70.4, 80.7) & 75.3 (69.8, 79.7) & 74.6 (70.1, 80.3) & 0.996 \\
Education (years)       & 16.0 (13.2, 18.0) & 16.0 (14.0, 18.0) & 16.0 (14.0, 18.0) & 0.552 \\
MMSE                    & 28.0 (26.0, 29.0) & 27.0 (25.0, 28.0) & 27.0 (26.0, 29.0) & $<0.001$ \\
CDR-SB                  & 1.5 (1.0, 1.5)    & 2.0 (1.0, 2.5)    & 1.5 (1.0, 2.0)    & $<0.001$ \\
FAQ                     & 1.0 (0.0, 3.0)    & 4.0 (1.0, 8.0)    & 2.0 (0.0, 6.0)    & $<0.001$ \\
MoCA                    & 26.0 (23.0, 27.0) & 26.0 (22.0, 27.0) & 26.0 (22.0, 27.0) & 0.556 \\
NPI                     & 1.0 (0.0, 2.0)    & 1.0 (0.0, 3.0)    & 1.0 (0.0, 3.0)    & 0.009 \\
ADAS-Cog 13             & 15.3 (10.5, 19.7) & 21.0 (17.7, 24.0) & 18.7 (14.3, 23.0) & $<0.001$ \\
Gender                  &                   &                   &                   & 0.172 \\
\quad 0.0               & 92 (68.7\%)       & 94 (60.3\%)       & 186 (64.1\%)      & \\
\quad 1.0               & 42 (31.3\%)       & 62 (39.7\%)       & 104 (35.9\%)      & \\
APOE4 Count             &                   &                   &                   & $<0.001$ \\
\quad 0                 & 87 (60.4\%)       & 56 (32.9\%)       & 143 (45.5\%)      & \\
\quad 1                 & 49 (34.0\%)       & 86 (50.6\%)       & 135 (43.0\%)      & \\
\quad 2                 & 8 (5.6\%)         & 28 (16.5\%)       & 36 (11.5\%)       & \\
\botrule
\end{tabular}
\end{table*}

\section{Hyperparameter optimization and computational environment}\label{secA2}
\subsection{Hardware and software environment}
All predictive modeling and hyperparameter optimization tasks were executed on a High-Performance Computing (HPC) cluster managed via the SLURM workload manager. Each array task was allocated a dedicated computational node with 16 CPU cores and 60 GB of RAM, with a maximum wall-clock time of 72 hours. To ensure strict reproducibility across all experiments, the software environment was fully containerized using a SquashFS image containing Python 3 and all necessary machine learning libraries. The asynchronous pipeline was executed identically for the multiple sclerosis, Parkinson's disease, ADNI, and synthetic cohorts.

\subsection{Hyperparameter search space}
During the independent randomized grid search cross-validation (Stage 2 of the ensemble pipeline), each base classifier was optimized to maximize the Area Under the Precision-Recall Curve (AUPRC). A maximum of $n_{\text{iter}}=20$ configurations were sampled per classifier during each of the 30 bootstrap iterations. The hyperparameter search spaces utilized for the random sampling are detailed in Table~\ref{tab_hyperparameters}.

\begin{table*}[t]
\caption{\textbf{Hyperparameter search space for the heterogeneous ensemble base classifiers.} Categorical options are listed in braces; uniform and log-uniform continuous distributions are denoted by $\mathcal{U}$ and $\log\mathcal{U}$, respectively; discrete integer ranges are denoted by $[min, max]$.}\label{tab_hyperparameters}
\centering
\begin{tabular}{l l}
\toprule
Classifier \& Hyperparameter & Search Space / Distribution \\
\midrule
\textbf{Random Forest} & \\
\quad \texttt{n\_estimators} & $[100, 1000]$ (step: 5) \\
\quad \texttt{max\_depth} & $[1, 110]$ \\
\quad \texttt{min\_samples\_split} & $[2, 30]$ \\
\quad \texttt{max\_features} & $\{\text{log2}, \text{sqrt}\}$ \\
\quad \texttt{class\_weight} & $\{\text{balanced}, \text{balanced\_subsample}\}$ \\
\midrule
\textbf{Extra Trees} & \\
\quad \texttt{n\_estimators} & $[1, 800]$ (step: 5) \\
\quad \texttt{max\_depth} & $[1, 40]$ \\
\quad \texttt{max\_features} & $[1, 40]$ \\
\quad \texttt{class\_weight} & $\{\text{balanced}, \text{balanced\_subsample}\}$ \\
\midrule
\textbf{XGBoost} & \\
\quad \texttt{n\_estimators} & $[50, 250]$ \\
\quad \texttt{learning\_rate} & $\mathcal{U}(0.01, 0.1)$ \\
\quad \texttt{max\_depth} & $[1, 20]$ \\
\quad \texttt{subsample} & $\mathcal{U}(0.4, 1.0)$ \\
\quad \texttt{gamma} & $\mathcal{U}(0, 0.5)$ \\
\quad \texttt{scale\_pos\_weight} & $\{1, 2, 3, 4, 5, 6, 7, 8, 9\}$ \\
\midrule
\textbf{Logistic Regression} & \\
\quad \texttt{C} & $\log\mathcal{U}(0.001, 1000)$ \\
\quad \texttt{max\_iter} & $[500, 1000]$ \\
\quad \texttt{fit\_intercept} & $\{\text{True}, \text{False}\}$ \\
\quad \texttt{class\_weight} & $\{\text{balanced}\}$ \\
\midrule
\textbf{Support Vector Classifier} & \\
\quad \texttt{C} & $\log\mathcal{U}(0.0001, 10000)$ \\
\quad \texttt{kernel} & $\{\text{linear}, \text{poly}, \text{rbf}, \text{sigmoid}\}$ \\
\quad \texttt{degree} & $[2, 4]$ \\
\quad \texttt{gamma} & $\mathcal{U}(0.001, 1.0)$ \\
\quad \texttt{coef0} & $\{0, 0.5, 1\}$ \\
\quad \texttt{shrinking} & $\{\text{True}, \text{False}\}$ \\
\quad \texttt{class\_weight} & $\{\text{balanced}\}$ \\
\botrule
\end{tabular}
\end{table*}

\section{Results on synthetic data}\label{secA3}
\subsection{Phase diagram of standard UQ failure (n=300 scenario).}
\begin{figure}[htpb]
    \centering
    \includegraphics[width=1\linewidth]{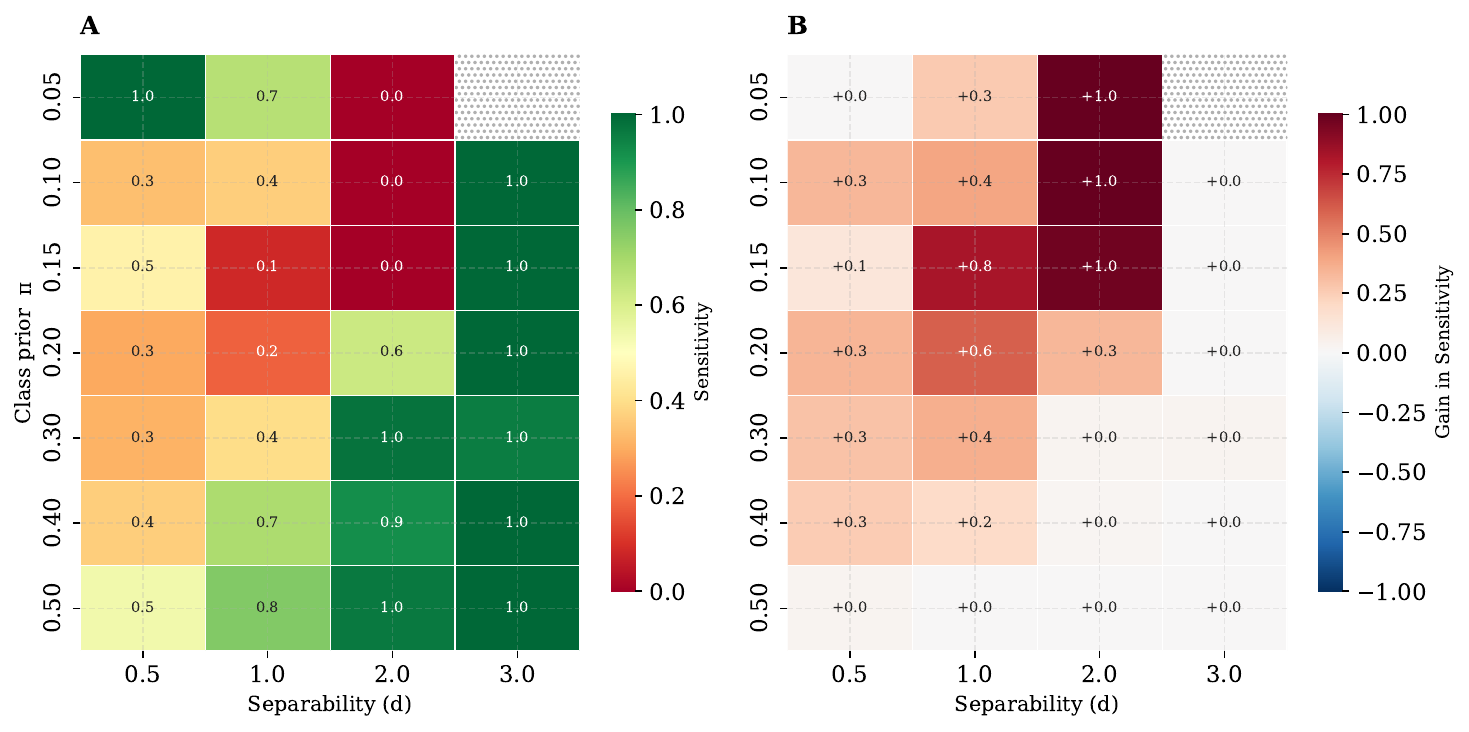}
    \caption{\textbf{Failure landscape of selective prediction under class imbalance.} Heatmaps illustrate minority class Sensitivity at a fixed 30\% global rejection rate across 56 synthetic conditions, parameterized by class prior ($\pi \in [0.05, 0.50]$, x-axis) and class separability ($d \in [0.5, 3.0]$, y-axis). \textbf{(A)} Standard global $H_{\text{Total}}$ exhibits a severe structural failure boundary. Under conditions of severe imbalance and overlapping distributions ($d \leq 1.0, \pi \leq 0.20$) the baseline actively defers minority patients (Sensitivity $\approx 0$). \textbf{(B)} The proposed $H_\tau$~+~CCR dual-architecture successfully rescues minority coverage within this diagnostic failure zone. Crucially, the architectural correction scales proportionally with the degree of imbalance, smoothly converging to identical standard baseline behavior when the natural prior is perfectly balanced ($\pi=0.50$).}
    \label{fig:fig_C1}
\end{figure}

Standard UQ failure phase map is reported in Figure \ref{fig:fig_C1} for $n=300$ total samples.

\subsection{Sensitivity stability scaling law (n=300 scenario).}
\begin{figure}[htpb]
    \centering
    \includegraphics[width=1\linewidth]{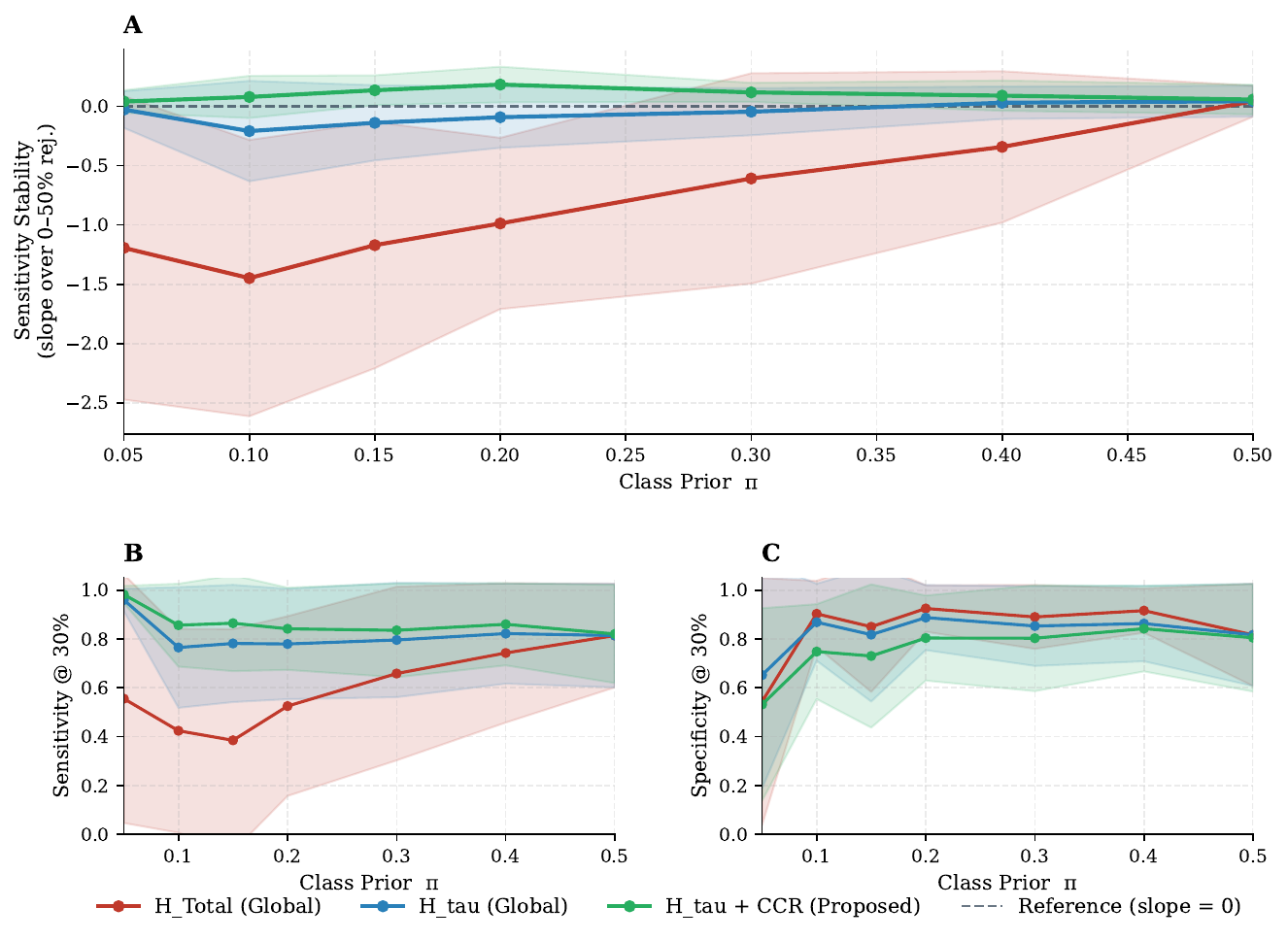}
    \caption{\textbf{Scaling law of minority Sensitivity under dynamic rejection across class priors (n=300).} 
    \textbf{(A)} Sensitivity stability, defined as the linear slope of minority Sensitivity across the $0\%\text{--}50\%$ rejection range. Standard global $H_{\text{Total}}$ (red) exhibits a severe non-linear scaling law, plunging into active minority exclusion as class imbalance increases. The proposed $H_\tau$~+~CCR dual-architecture (green) fully neutralizes this artifact, maintaining a safe trajectory near zero across all priors. 
    \textbf{(B)} Sensitivity evaluated at a fixed 30\% global rejection rate. Standard UQ systematically collapses under severe imbalance ($\pi \leq 0.20$), whereas the proposed framework guarantees robust minority retention. 
    \textbf{(C)} Specificity (majority class retention) evaluated at the exact same 30\% rejection rate. Under a fixed global rejection constraint, standard UQ maximizes Specificity by disproportionately concentrating deferrals within the minority class. By recovering minority Sensitivity (Panel B), the proposed framework mathematically redistributes these deferrals, resulting in a balanced and class-proportional rejection profile.}
    \label{fig:fig_C2}
\end{figure}

The stability of the Sensitivity computed in the range $0\%-50\%$ together with the Sensitivity and Specificity at a fixed rejection rate of 30\%, for the scenario $n=300$ samples, is reported in Figure \ref{fig:fig_C2}.

\subsection{Rejection curves on synthetic data.}
\begin{figure}[htpb]
    \centering
    \includegraphics[width=1\linewidth]{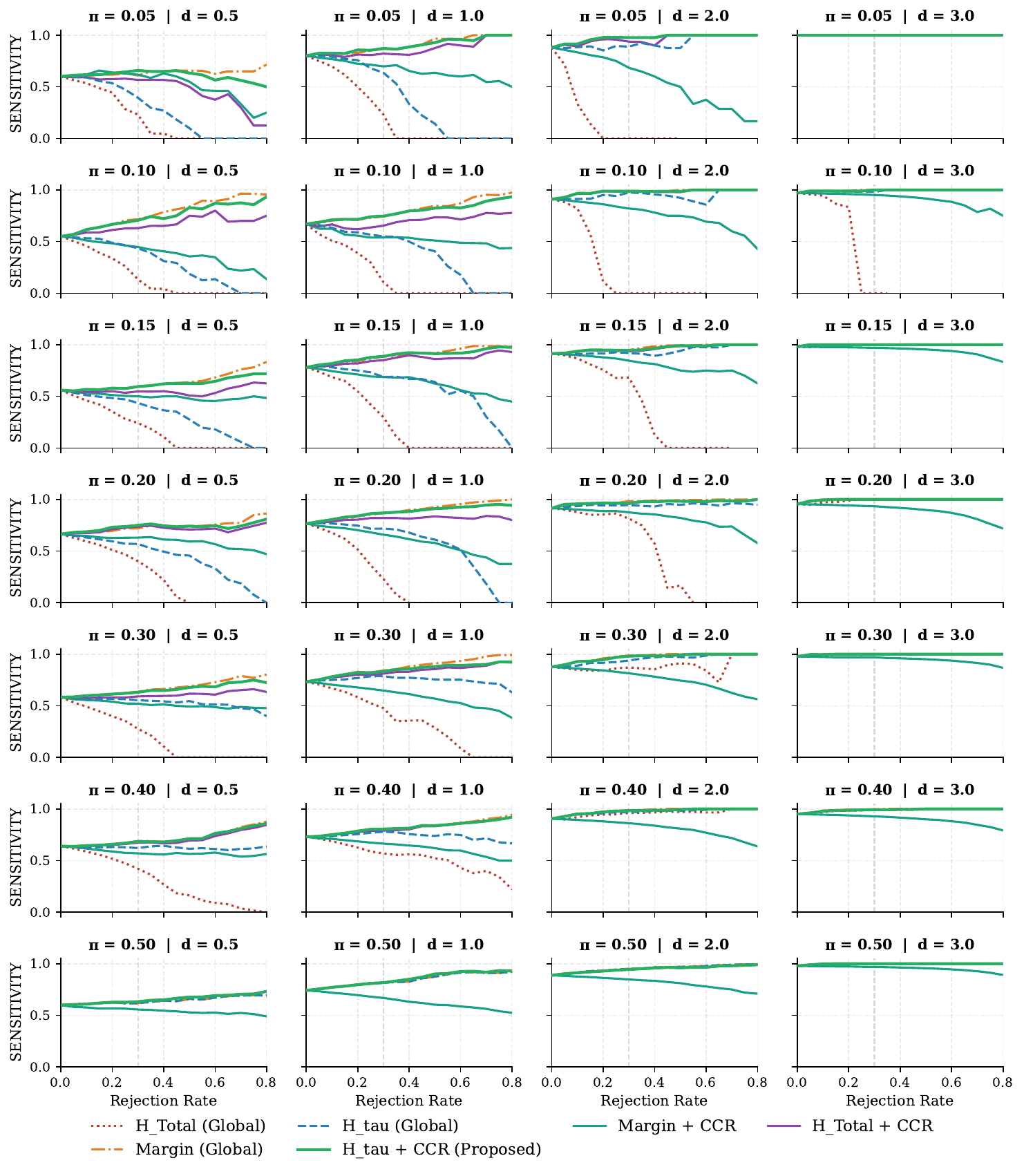}
    \caption{\textbf{Rejection curves on synthetic data ($n=1000$).} Minority class Sensitivity is evaluated across a grid of class priors ($\pi$) and separability ($d$), comparing the proposed framework against global baselines.}
    \label{fig:figA1_rej_curves_1000}
\end{figure}

\begin{figure}[htpb]
    \centering
    \includegraphics[width=1\linewidth]{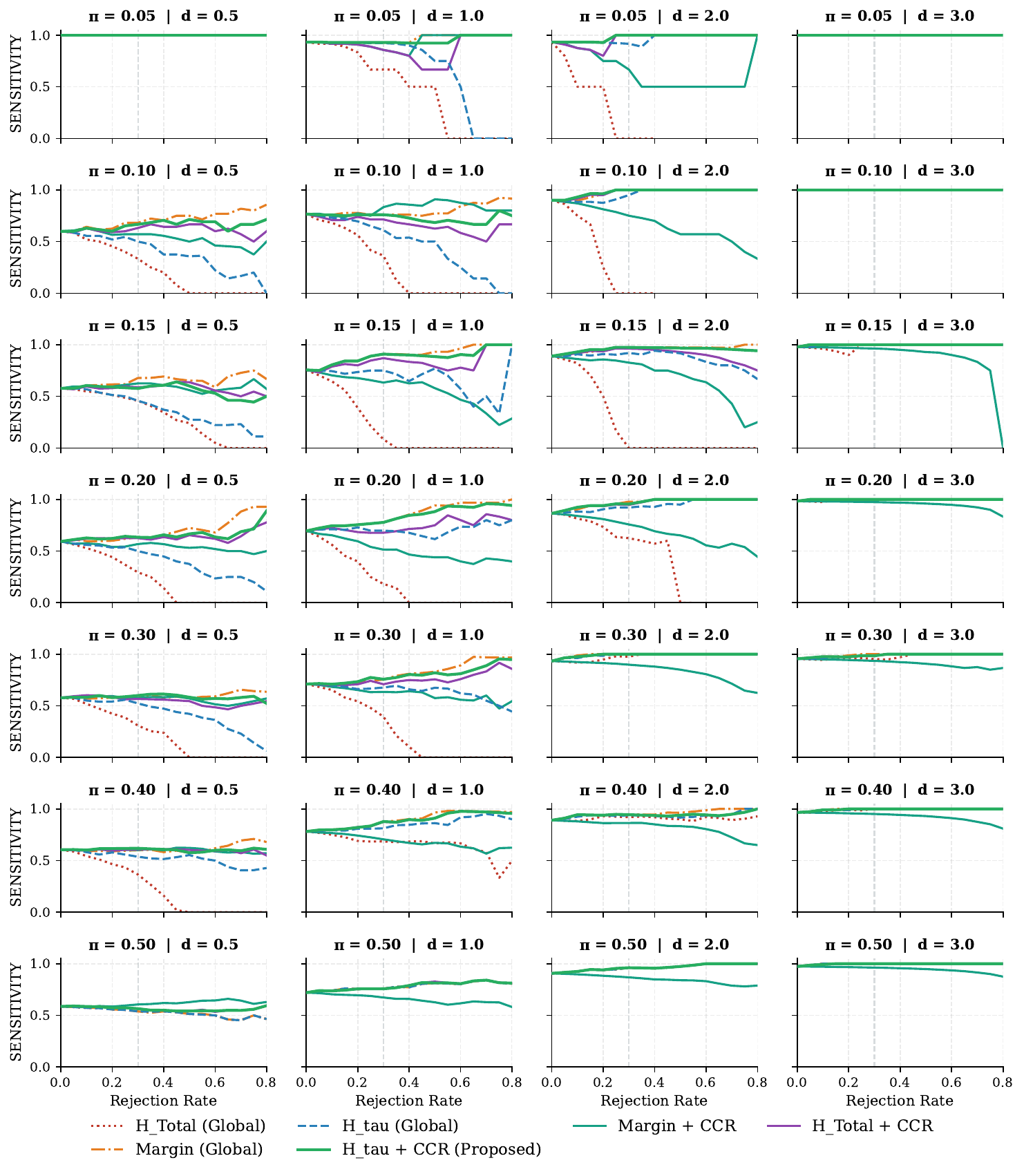}
    \caption{\textbf{Rejection curves on synthetic data ($n=300$).} Minority class Sensitivity is evaluated across a grid of class priors ($\pi$) and separability ($d$), comparing the proposed framework against global baselines.}
    \label{fig:figA1_rej_curves_300}
\end{figure}

To fully capture the dynamic degradation of minority coverage, Figures \ref{fig:figA1_rej_curves_1000} and \ref{fig:figA1_rej_curves_300} map the complete rejection trajectories across all 56 synthetic conditions for $N=1000$ and $N=300$, respectively. These panels track minority class Sensitivity as the rejection rate increases continuously from 0\% to 80\%. The curves visually confirm the structural failure boundaries identified in the phase diagrams: under conditions of severe imbalance and overlapping distributions ($\pi \le 0.20$, $d \le 1.0$), standard global metrics ($H_{\text{Total}}$) trigger a rapid and complete Sensitivity collapse early in the triage process. While applying the isolated $H_{\tau}$ signal globally delays this depletion by resolving the entropy mismatch, it ultimately fails at higher rejection rates due to uncorrected spatial penalties. In contrast, the proposed $H_{\tau}$~+~CCR architecture consistently maintains stable, bounded minority protection across the entire rejection continuum, smoothly converging to standard baseline behavior only as the clinical prior approaches perfect balance ($\pi = 0.50$).


\subsection{Controlled evaluation of structural uncertainty divergence}
To isolate the effect of class prevalence on uncertainty estimation, we replicated the asymmetry test using strictly controlled synthetic datasets. This framework allows us to evaluate the structural divergence of ensemble disagreement independent of the varying feature spaces and biological complexities present in real-world clinical cohorts.

\begin{figure}[htpb]
    \centering
    \includegraphics[width=1\linewidth]{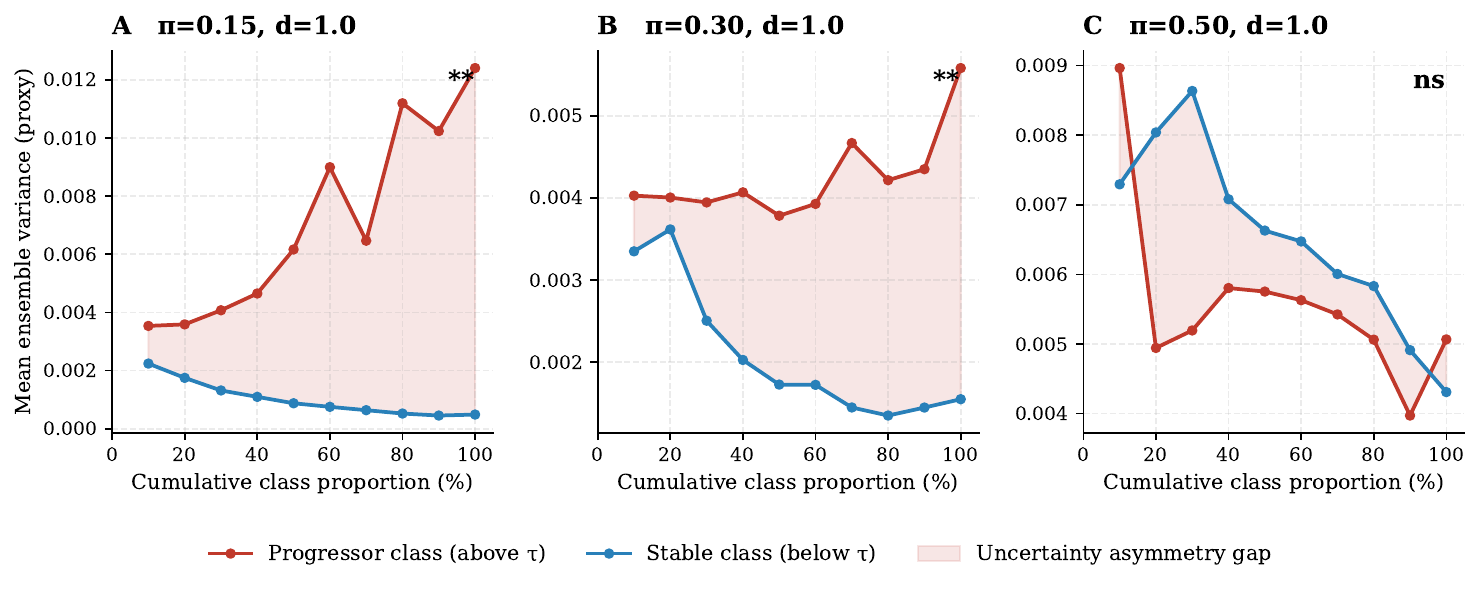}
    \caption{\textbf{Structural uncertainty divergence at a constrained sample size ($N=300$).} Evaluation of ensemble variance across cumulative class proportions in synthetic datasets. At a high class imbalance ($\pi=0.15$), a severe and statistically significant uncertainty gap emerges between the classes ($p < 0.01$). This demonstrates that even when minority instances are correctly identified with high probability, they suffer from structurally higher ensemble variance simply due to the epistemic sparsity of the local region. This divergence narrows at moderate imbalance ($\pi=0.30$, $p < 0.01$) and fully dissolves under balanced conditions ($\pi=0.50$, $p = \text{n.s.}$), mirroring the real-world clinical vulnerabilities.}
    \label{fig:figC4_uncertainty_div_n300}
\end{figure}

\begin{figure}[htpb]
    \centering
    \includegraphics[width=1\linewidth]{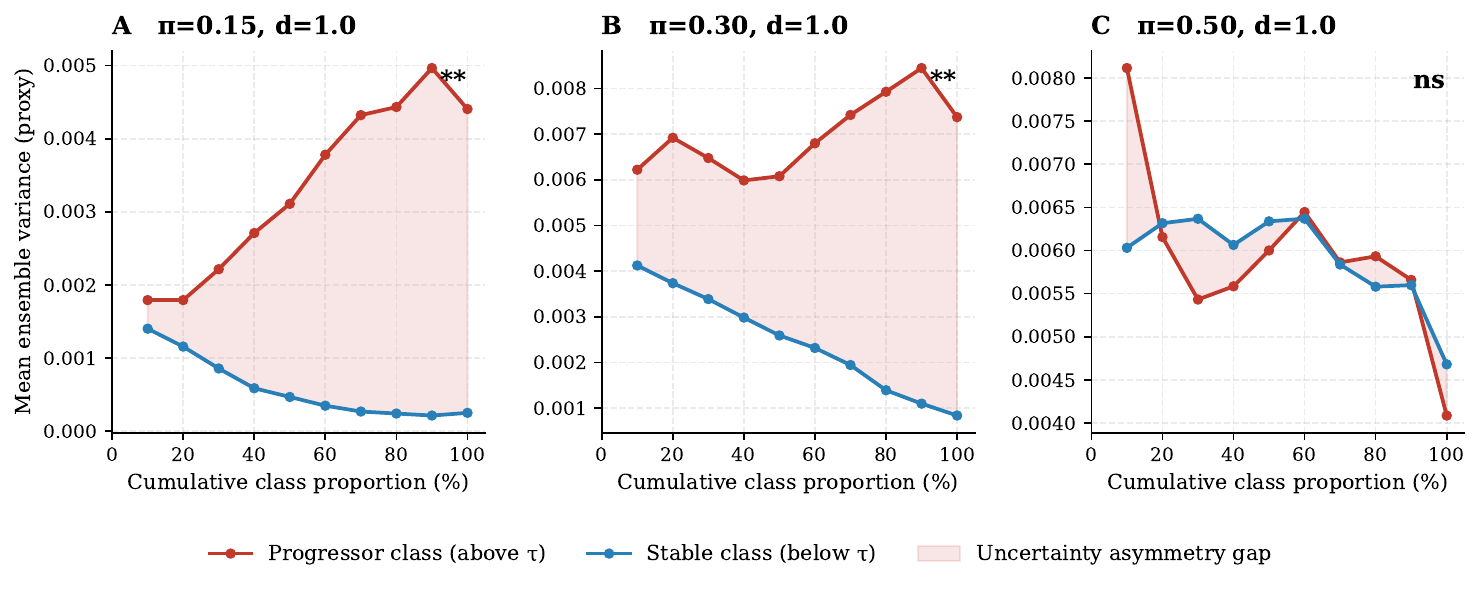}
    \caption{\textbf{Structural uncertainty divergence at a large sample size ($N=1000$).} Replication of the asymmetry test with an expanded overall cohort ($N=1000$). While the absolute magnitude of the ensemble variance decreases globally compared to $N=300$, the relative structural divergence remains firmly intact. The uncertainty asymmetry gap persists significantly at $\pi=0.15$ ($p < 0.01$) and $\pi=0.30$ ($p < 0.01$), only dissipating when parity ($\pi=0.50$, $p = \text{n.s.}$) is achieved.}
    \label{fig:figC4_uncertainty_div_n1000}
\end{figure}

For these experiments, we fixed the underlying class separability (distance between class means) at $d=1.0$ and systematically varied the minority class prior ($\pi \in \{0.15, 0.30, 0.50\}$). To evaluate how absolute sample sparsity interacts with relative imbalance, we conducted the analysis at two different overall dataset sizes: $N=300$ in Figure \ref{fig:figC4_uncertainty_div_n300} and $N=1000$ in Figure \ref{fig:figC4_uncertainty_div_n1000}.

\section{Results on real-world data}\label{secA4}

\begin{figure}[htpb]
    \centering
    \includegraphics[width=1\linewidth]{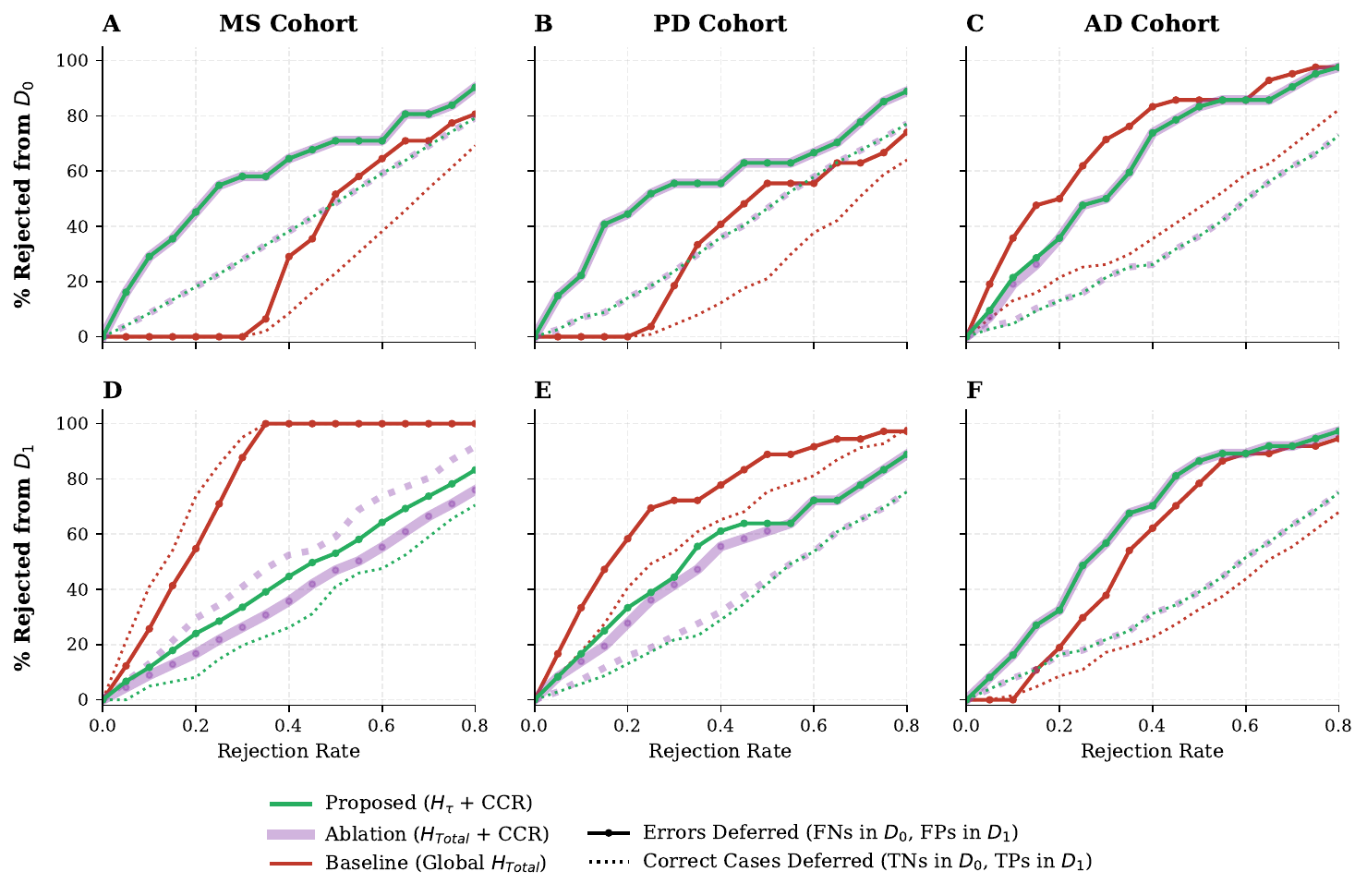}
    \caption{\textbf{Continuous error-filtering dynamics within the isolated triage queues.} The plots trace the percentage of boundary errors (solid lines) versus confident correct predictions (dotted lines) deferred from the $\mathcal{D}_0$ queue (top row) and $\mathcal{D}_1$ queue (bottom row). By isolating the queues, the proposed $H_\tau$~+~CCR architecture (green) forces monotonic uncertainty metrics to aggressively target boundary errors. Consequently, the framework symmetrically rejects true errors (FNs and FPs) at substantially higher rates than correct predictions. The ablation metric ($H_{\text{Total}}$~+~CCR, purple) mimics the proposed framework in the $\mathcal{D}_0$ queue due to its monotonic behavior below the threshold, but fails in the $\mathcal{D}_1$ queue where standard entropy peaks at 0.5 (deferring true positives faster than false positives). This confirms that while queue isolation (CCR) is sufficient for filtering false negatives, threshold-aware geometry ($H_\tau$) is strictly required to filter false positives.}
    \label{fig:figD1_reject_dynamics}
\end{figure}

\begin{table*}[htbp]
\caption{\textbf{Generalized risk-coverage evaluation computed over a strict common support.} This table recalculates the class-conditioned Area Under the Generalized Risk-Coverage Curve (ccAUGRC) and Global AUGRC using a standardized integration domain across all methods. The class disparity column has been added to mirror Table 2. Values are point estimates with 95\% Confidence Intervals derived from 1000 post-prediction bootstrap iterations. Significance testing compares each baseline to the proposed reference framework ($H_\tau$~+~CCR), with $p$-values adjusted for multiple comparisons using the Holm-Bonferroni method. Lower values indicate better safety (lower risk).}\label{tab_risk_results_cs}
\centering
\resizebox{\textwidth}{!}{
\begin{tabular}{l l l l l l}
\toprule
Dataset & Uncertainty Framework & Progressor ccAUGRC (95\% CI) & Stable ccAUGRC (95\% CI) & Class Disparity & Global AUGRC (95\% CI) \\
\midrule
\multirow{6}{*}{\textbf{MS} ($\tau=\pi=0.13$)} & $H_{\text{Total}}$ (Global) & 0.2044 [0.1300, 0.2863]$^{\uparrow\textcolor{red}{***}}$ & 0.0413 [0.0317, 0.0518]$^{\downarrow\textcolor{red}{***}}$ & 0.1631 & 0.0679 [0.0556, 0.0813] \\
  & Margin (Global) & 0.0931 [0.0562, 0.1378]$^{\downarrow\text{ns}}$ & 0.1463 [0.1247, 0.1672]$^{\uparrow\textcolor{red}{***}}$ & 0.0532 & 0.1430 [0.1233, 0.1623] \\
  & $H_\tau$ (Global) & 0.1693 [0.1100, 0.2331]$^{\uparrow\textcolor{red}{***}}$ & 0.0561 [0.0449, 0.0680]$^{\downarrow\textcolor{red}{***}}$ & 0.1132 & 0.0734 [0.0605, 0.0867] \\
  & Random~+~CCR & 0.1418 [0.0914, 0.1953]$^{\uparrow\textcolor{red}{***}}$ & 0.1362 [0.1192, 0.1525]$^{\uparrow\textcolor{red}{**}}$ & 0.0056 & 0.1404 [0.1231, 0.1577] \\
  & $H_{\text{Total}}$~+~CCR & 0.1241 [0.0773, 0.1741]$^{\uparrow\textcolor{red}{**}}$ & 0.1392 [0.1223, 0.1558]$^{\uparrow\textcolor{red}{*}}$ & 0.0151 & 0.1404 [0.1235, 0.1577] \\
  & \textbf{$H_\tau$~+~CCR} & \textbf{0.1025 [0.0614, 0.1486]} & \textbf{0.1288 [0.1123, 0.1454]} & \textbf{0.0263} & \textbf{0.1275 [0.1119, 0.1442]} \\
\midrule
\multirow{6}{*}{\textbf{PD} ($\tau=\pi=0.39$)} & $H_{\text{Total}}$ (Global) & 0.1609 [0.1060, 0.2169]$^{\uparrow\textcolor{red}{***}}$ & 0.0431 [0.0218, 0.0660]$^{\downarrow\textcolor{red}{***}}$ & 0.1178 & 0.0918 [0.0646, 0.1191] \\
  & Margin (Global) & 0.0840 [0.0454, 0.1275]$^{\downarrow*}$ & 0.0946 [0.0603, 0.1311]$^{\uparrow\text{ns}}$ & 0.0106 & 0.0973 [0.0688, 0.1255] \\
  & $H_\tau$ (Global) & 0.1063 [0.0602, 0.1556]$^{\uparrow*}$ & 0.0756 [0.0463, 0.1095]$^{\downarrow*}$ & 0.0307 & 0.0926 [0.0649, 0.1188] \\
  & Random~+~CCR & 0.1244 [0.0808, 0.1740]$^{\uparrow*}$ & 0.1106 [0.0768, 0.1466]$^{\uparrow\textcolor{red}{*}}$ & 0.0138 & 0.1228 [0.0940, 0.1541] \\
  & $H_{\text{Total}}$~+~CCR & 0.0985 [0.0556, 0.1448]$^{\uparrow\text{ns}}$ & 0.0922 [0.0597, 0.1273]$^{\uparrow\text{ns}}$ & 0.0063 & 0.0987 [0.0720, 0.1267] \\
  & \textbf{$H_\tau$~+~CCR} & \textbf{0.0965 [0.0535, 0.1401]} & \textbf{0.0885 [0.0554, 0.1223]} & \textbf{0.0080} & \textbf{0.0958 [0.0681, 0.1227]} \\
\midrule
\multirow{6}{*}{\textbf{AD} ($\tau=\pi=0.54$)} & $H_{\text{Total}}$ (Global) & 0.0567 [0.0352, 0.0834]$^{\downarrow\textcolor{red}{***}}$ & 0.0989 [0.0669, 0.1358]$^{\uparrow\textcolor{red}{***}}$ & 0.0422 & 0.0756 [0.0570, 0.0966] \\
  & Margin (Global) & 0.0781 [0.0519, 0.1085]$^{\uparrow\text{ns}}$ & 0.0733 [0.0459, 0.1057]$^{\downarrow\text{ns}}$ & 0.0048 & 0.0763 [0.0572, 0.0966] \\
  & $H_\tau$ (Global) & 0.0706 [0.0460, 0.0996]$^{\downarrow\text{ns}}$ & 0.0808 [0.0514, 0.1151]$^{\uparrow\text{ns}}$ & 0.0102 & 0.0755 [0.0567, 0.0964] \\
  & Random~+~CCR & 0.1156 [0.0831, 0.1526]$^{\uparrow\textcolor{red}{***}}$ & 0.1203 [0.0830, 0.1587]$^{\uparrow\textcolor{red}{***}}$ & 0.0047 & 0.1203 [0.0940, 0.1479] \\
  & $H_{\text{Total}}$~+~CCR & 0.0767 [0.0517, 0.1062]$^{\uparrow\text{ns}}$ & 0.0751 [0.0471, 0.1062]$^{\uparrow\text{ns}}$ & 0.0016 & 0.0764 [0.0573, 0.0973] \\
  & \textbf{$H_\tau$~+~CCR} & \textbf{0.0761 [0.0514, 0.1057]} & \textbf{0.0744 [0.0462, 0.1054]} & \textbf{0.0017} & \textbf{0.0758 [0.0570, 0.0966]} \\
\bottomrule
\end{tabular}
}
\vspace{1ex}
{\raggedright \footnotesize
Significance vs. Proposed ($H_\tau$~+~CCR): *** $p<0.001$, ** $p<0.01$, * $p<0.05$, ns: not significant. \\
Arrows indicate if the baseline is structurally worse ($\uparrow$, higher risk) or better ($\downarrow$, lower risk). \\
Red markers ($\textcolor{red}{*}$) indicate that statistical significance is maintained after Holm-Bonferroni correction ($\alpha=0.05$). \par}
\end{table*}
\end{appendices}

\newpage
\bibliography{references}%

\end{document}